\documentclass{article}

\usepackage{microtype}
\usepackage{graphicx}
\usepackage{subfigure}
\usepackage{booktabs} 

\usepackage{amsmath}
\usepackage{bm}
\usepackage{amsfonts} 
\usepackage{dsfont}
\DeclareMathOperator*{\argmax}{argmax}

\usepackage{multirow}
\usepackage{subcaption}
\usepackage{caption}
\usepackage{acronym}
\usepackage[table,xcdraw]{xcolor}
\usepackage{tikz}
\usetikzlibrary{arrows.meta}

\usepackage{hyperref}

\usepackage{eqparbox}

\usepackage[accepted]{icml2025}

\usepackage{amsmath}
\usepackage{amssymb}
\usepackage{mathtools}
\usepackage{amsthm}

\usepackage[capitalize,noabbrev]{cleveref}

\theoremstyle{plain}
\newtheorem{theorem}{Theorem}[section]

\theoremstyle{definition}

\theoremstyle{remark}

\acrodef{BO}{Bayesian Optimisation}
\acrodef{GP}{Gaussian Process}
\acrodefplural{GP}[GPs]{Gaussian Processes}
\acrodef{LSBO}{Latent Space \ac{BO}}
\acrodef{VAE}{Variational AutoEncoder}
\acrodef{GAN}{Generative Adversarial Network}
\acrodef{COWBOYS}{Categorical Optimisation With Belief Of underlYing Structure}
\acrodef{MCMC}{Markov Chain Monte-Carlo}
\acrodef{PCN}{Preconditioned Crank-Nicolson}
\acrodef{SMILES}{Simplified Molecular Input Line Entry System}
\acrodef{UCB}{Upper Confidence Bounds}
\acrodef{EI}{Expected Improvement}
\acrodef{PI}{Probability of Improvement}
\acrodef{NN}{Neural Network}
\acrodef{LOLBO}{LOcal Latent \ac{BO}}
\acrodef{ARD}{Automatic Relevance Determination}
\acrodef{ECFP}{Extended Connectivity Fingerprints}
\acrodef{FCFP}{Functional Class Fingerprints}
\acrodef{MACCS}{Molecular Access System}
\acrodef{SLERP}{spherical linear interpolation}

\usepackage[textsize=tiny]{todonotes}

\icmltitlerunning{Return of the Latent Space COWBOYS}

\begin{document}

\twocolumn[
\icmltitle{Return of the Latent Space COWBOYS: Re-thinking the use of VAEs for Bayesian Optimisation of Structured Spaces}



\icmlsetsymbol{equal}{*}

\begin{icmlauthorlist}
\icmlauthor{Henry B. Moss}{hm1,hm2}
\icmlauthor{Sebastian W. Ober}{az1}
\icmlauthor{Tom Diethe}{az2}
\end{icmlauthorlist}

\icmlaffiliation{hm1}{School of Mathematical Sciences, Lancaster University, UK}
\icmlaffiliation{hm2}{Department of Applied Maths and Theoretical Physics, University of Cambridge, UK}

\icmlaffiliation{az1}{Oncology R\&D, AstraZeneca, Gaithersburg, USA 
}
\icmlaffiliation{az2}{ Biopharma R\&D, AstraZeneca, Cambridge, UK
}

\icmlcorrespondingauthor{Henry B. Moss}{hm493@cam.ac.uk}

\icmlkeywords{Machine Learning, ICML}

\vskip 0.3in
]



\printAffiliationsAndNotice{}  


\begin{abstract}
Bayesian optimisation in the latent space of a \acf{VAE} is a powerful framework for optimisation tasks over complex structured domains, such as the space of scientifically interesting molecules. However, existing approaches tightly couple the surrogate and generative models, which can lead to suboptimal performance when the latent space is not tailored to specific tasks, which in turn has led to the proposal of increasingly sophisticated algorithms. In this work, we explore a new direction, instead proposing a decoupled approach that trains a generative model and a \acf{GP} surrogate separately, then combines them via a simple yet principled Bayesian update rule. This separation allows each component to focus on its strengths— structure generation from the \ac{VAE} and predictive modelling by the \ac{GP}. We show that our decoupled approach improves our ability to identify high-potential candidates in molecular optimisation problems under constrained evaluation budgets.
\end{abstract}

\section{Introduction}\label{sec:introduction}

First introduced by \citet{gomez2018automatic}, \acf{BO} over latent spaces has emerged as a powerful technique for optimising over structures. Rather than performing challenging combinatorial or high-dimensional optimisation directly on discrete structures --- such as molecules or proteins --- \acf{LSBO} first maps inputs into a fixed-dimensional Euclidean latent space, where standard surrogate models and gradient-based acquisition routines can be employed. Candidate points selected by the optimiser are then decoded back into the original structured domain to yield new query points.

Learning a complex mapping from variable-size structured inputs to fixed-size Euclidean representations requires more data than is typically gathered in LSBO itself. Therefore, \ac{LSBO} relies on embeddings pre-trained on a related task with abundant unlabelled data (e.g., using a \ac{VAE} latent space trained on a large set of valid molecules). However, it is well known \citep{chuinversion} that fitting surrogates in the latent space of a \ac{VAE} can lead to suboptimal modelling; therefore, the recent \ac{LSBO} literature has focused on introducing increasingly sophisticated heuristics to fine-tune the VAE on newly acquired experimental data. These approaches remain challenged by the persistent risk of overfitting when training neural network components of Bayesian models on limited datasets \cite{ober2021promises}. In this work, we also make a new argument that the prevalent approach of limiting the search to a fixed box subset of the latent space --- necessary to apply traditional \ac{BO} methods --- also limits the effectiveness  of \ac{LSBO}.

\begin{figure}
\centering
\includegraphics[width=0.47\textwidth]{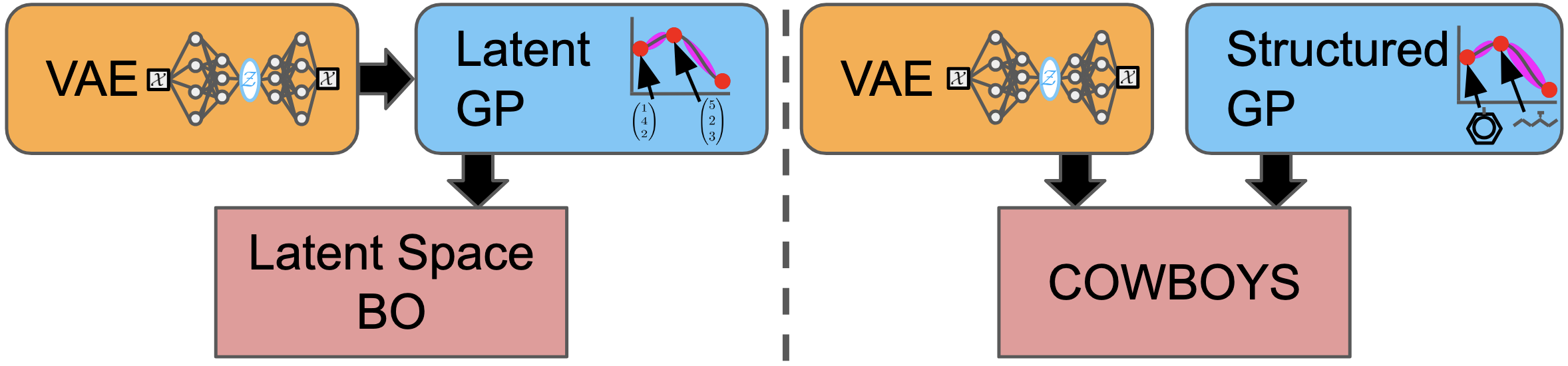}
    \caption{Unlike \ac{LSBO} where \acp{GP} are fit in a VAE's latent space, \ac{COWBOYS}'s \ac{GP} is fit in structure space, decoupled from its \ac{VAE}.} 
    \label{fig:summary}
    \vspace{-1cm}
\end{figure}

While the \ac{VAE} is designed purely as a generative model, \ac{LSBO} attempts to repurpose it for discrimination --- a fundamental mismatch that gives rise to the pathologies discussed above. We propose an alternative approach that preserves the original motivation of the \ac{VAE} as a generative model
by coupling it with a separately trained surrogate \ac{GP} model via a novel Bayesian formulation. Our primary goal is straightforward: \textit{develop a sampling strategy that increasingly favours structures with desirable objective values.}

We propose \acf{COWBOYS}, a novel principled framework that unifies \acp{GP} and \acp{VAE} within a \ac{BO} loop. Although we focus on molecular search—the most popular application of \ac{LSBO}—where we demonstrate that \ac{COWBOYS} excels under heavily constrained optimisation budgets, our approach is general and applicable to any structured problem for which a suitable structural GP kernel can be defined (see our discussion in Section \ref{sec:discussion}).

\section{Background} \label{sec:background}

\subsection{Bayesian Optimisation} 
\label{sec:bo_intro}

\acf{BO} \cite{mockus2005bayesian} is a powerful framework for the efficient optimisation of costly black-box functions $f : \mathcal{X} \rightarrow \mathds{R}$.
For a successful application of \ac{BO}, we typically need three components: a search space, a surrogate model, and an acquisition function.
Given some initial data, 
\ac{BO} constructs a surrogate model of the data, which is then used in tandem with an acquisition function to determine which point in the search space will be most valuable to acquire. Once this point is chosen, we query the black-box function at this point and update the surrogate model, repeating until the evaluation budget is exhausted. 

\textbf{Search spaces}. In order for optimisation to succeed, the space over which we attempt to find the optimal value must be suitably restricted.
For instance, when dealing with a Euclidean space $\mathds{R}^d$, it is typical to restrict the search space to a box region $[-\delta, \delta]^d$, where $\delta$ is typically chosen using some prior knowledge about the expected location of the optimum \cite{hvarfner2022pibo}.
In this example, choosing too large a $\delta$ will lead to slow convergence due to the need for more data so that the surrogate can be effective, whereas too small a $\delta$ will risk missing the optimum. 
Later, we will discuss how choosing appropriate search spaces is a key bottleneck for \ac{LSBO} and how 
\ac{COWBOYS} addresses this.

\textbf{Surrogate models}. BO requires surrogate models that accurately quantify uncertainty while retaining flexibility: as such, \acfp{GP} \cite{rasmussen2003gaussian} are a standard choice.
A \ac{GP} can be defined as an infinite collection of random variables, any finite number of which are Gaussian distributed, and is entirely defined by a mean function $\mu(\cdot)$ and a kernel $k(\cdot,\cdot)$.
Under a \ac{GP} $f \sim \mathcal{GP}(\mu, k)$, 
making a prediction $\hat{f}$ at a test point $\hat{\bm{x}}$ amounts to conditioning on the observed data $D$ : $p(\hat{f} | D) = \mathcal{N}(\mu^*, \Sigma^*)$, where $\hat{\mu}$ and $\hat{\Sigma}$ can be computed in closed form via the properties of Gaussians with $O(N^3)$ computational complexity and $O(N^2)$ memory.

\textbf{Acquisition functions}. Acquisition functions  measure the utility of acquiring an unseen point $\bm{x}$ according to the predictions of our surrogate model. Common acquisition functions measure the \ac{EI}, \ac{PI} \cite{bergstra2011algorithms} or \ac{UCB} \cite{auer2000using} of candidate points, or use information theoretical arguments \cite{hennig2012entropy, moss2021gibbon}, e.g.,
\begin{align}
    \alpha_{PI}(\bm{x}; D) = \mathds{P}(f(\bm{x}) > f^* | D), \label{eq:PI}
\end{align}
where $f^*$ is the best objective value observed so far.

\textbf{Covariance (kernel) functions}. In this work, where we must build a surrogate model over the space of molecules, we use the Tanimoto kernel \citep{tripp2023tanimoto}:
\begin{align}
    K_T(\bm{m}, \bm{m}') = \sigma^2\frac{\bm{m} \cdot \bm{m}'}{\|\bm{m}\|^2 + \|\bm{m}'\|^2 - \bm{m} \cdot \bm{m}'}, 
    \nonumber
\end{align}
where $\bm{m}$ and $\bm{m}'$ are molecular \emph{fingerprints} representation vectors of molecules and  $\sigma\in\mathds{R}^{+}$ is a tunable scaling factor. We follow the advice of \citet{tripp2024diagnosing} and use count-based  vectors that count specific molecular features \citep{landrum2013rdkit}. 
Structural kernels such as the Tanimoto kernel, but also string kernels \cite{moss2020boss} and graph kernels \cite{kriege2020survey}, can often outperform deep learning alternatives, particularly in low- to medium-data regimes \citep{moss2020gaussian,griffiths2022data}.




\subsection{Variational AutoEncoders} 
\acfp{VAE} \citep{kingma2014auto, rezende2014stochastic} are deep generative models that model data through a two-step process: sampling a latent variable $\bm{z}$ (with dimension smaller than $\mathcal{X}$) from a prior $p(\bm{z})$, followed by sampling $\bm{x}$ through a distribution $p_{\theta}(\bm{x}|\bm{z})$, resulting in a generative model $\bm{x} \sim p_\theta(\bm{x})$.
The latter of these steps is performed by a neural network, known as the \emph{decoder}, with parameters $\theta$.
The decoder outputs the parameters of the conditional distribution $p_{\theta}(\bm{x}|\bm{z})$, for instance the mean and variance of a normal distribution, or class probabilities for a categorical distribution.
In order to train a VAE, we require the posterior $p_{\theta}(\bm{z}|\bm{x})$. 
However, as this is intractable, we use a neural network \emph{encoder} with parameters $\phi$ to approximate it through amortised inference: $q_\phi(\bm{z}|\bm{x}) = \mathcal{N}(\bm{\mu}(\bm{x}),\bm{\sigma^2}(\bm{z})I)$, where $\phi$ parameterises $\bm{\mu}(\cdot)$ and $\bm{\sigma^2}(\cdot)$.
The encoder and decoder are trained jointly through a minibatch-friendly lower bound to the marginal likelihood of the data.


\section{The Pitfalls of Latent Space BO}

In what follows, we assume that we have access to a pre-trained \ac{VAE} with a $d$-dimensional latent space $\mathcal{Z}=\mathds{R}^d$, resulting in a decoding distribution $p_{\theta}(\textbf{x}|\textbf{z})$, and imply a data distribution $p_{\theta}(\bm x)$.

\subsection{Latent Space BO}
\acf{LSBO} \citep{gomez2018automatic} can be viewed as standard \ac{BO} conducted over an alternative search space --- the latent space $\mathcal{Z}$ of a \ac{VAE} --- and using a different surrogate model, one modelling the mapping $g:\mathcal{Z}\rightarrow\mathds{R}$ from latent codes to objective function values with a surrogate $\tilde{g}$. At the $n^{\textrm{th}}$ optimisation step, \ac{LSBO} proceeds analogously to standard BO. For example, under the PI acquisition function, \ac{LSBO} selects a new latent code $\bm z_n$ with highest utility
\begin{align}
\bm z_n\leftarrow\textcolor{black}{\argmax_{\bm z\in\mathcal{Z}}}\;\mathds{P}(\textcolor{black}{\tilde{g}(}\bm z\textcolor{blue}{)} > f^* | D^{\mathcal{Z}}_{n-1}),
\nonumber
\end{align}
where the surrogate $\tilde{g}$ is trained on the dataset $D^{\mathcal{Z}}_{n-1}= \{(\bm z_i, y_i)\}_{i=1}^{n-1}$ of latent code-evaluation pairs.
The result $z_n$ is then decoded and the resulting structure is evaluated on the objective function (see Algorithm \ref{alg:lsbo}).

\begin{algorithm}[tb]
   \caption{\label{alg:lsbo}Latent Space Bayesian Optimisation}
\begin{algorithmic}
   \STATE {\bfseries Input:} budget $N$, init size $N_{\textrm{init}}$, search bounds $\delta$
   \STATE Clip search space $\mathcal{Z}_{\delta}\leftarrow[-\delta,\delta]^d\subset\mathcal{Z}$
   \FOR{$n\in\{1,..,N\}$}
   \IF[initial design]{$n<N_{\textrm{init}}$}
   \STATE $\bm z_n \sim \textrm{SpaceFillingDesign}(\mathcal{Z}_{\delta})$ 
   \ELSE[sequential optimisation]
   \STATE $\bm z_n\leftarrow\argmax_{\bm z\in\mathcal{Z}_{\delta}}\;\alpha(\bm z; D_{n-1}^{\mathcal{Z}})$ \COMMENT{e.g. Eq. (\ref{eq:PI})}
   \ENDIF
   \STATE Decode chosen latent $\bm{x}_n\sim p_{\theta}(\bm x|\bm z_n)$
   \STATE Evaluate new molecular structure $y_n\leftarrow f(\bm{x}_n)$
    \STATE Update dataset $D^{\mathcal{Z}}_n\leftarrow D^{\mathcal{Z}}_{n-1} \bigcup \{(\bm z_n,y_n)\}$
    \STATE Fit latent space GP on $D^{\mathcal{Z}}_n$
   \ENDFOR
    \STATE \textbf{return} Believed optimum across $\{\bm{z}_1,..,\bm{z}_n\}$
\end{algorithmic}
\end{algorithm}


We dedicate the remainder of this section to precisely state the two primary pathological issues caused by this current way of coupling \acp{VAE} and \acp{GP} in \ac{LSBO}: i) fitting \acp{GP} in \ac{VAE} latent spaces can lead to poor surrogate models, and ii) it is challenging to define a search space within the latent space so that it reliably targets the most performant regions of the decoder and so leads to the selection and generation of high-quality and useful molecular structures (i.e., avoiding chemically invalid or biologically irrelevant molecules).

\subsection{LSBO Surrogate Models can be Poor Predictors}
\label{section:poor_model}

The premise behind the efficiency of BO is the ability to build an effective surrogate model by exploiting the smoothness of our objective $f$ over the search space $\mathcal{X}$. However, as demonstrated by many \citep[e.g.,][]{griffiths2020constrained,moss2020boss, tripp2020sample}, the mapping $g$ from the latent space to objective function values can be significantly more challenging to learn with a GP than $f$. Consequently, \ac{LSBO} often ends up with a poor surrogate model that under-represents variations, over-emphasises sub-optimal areas of the space, and fails to extrapolate the target property to not-yet-evaluated structures ---  a notion typically referred to by the catch-all term ``poor latent space alignment," generally attributed to two \ac{VAE} properties:

\textbf{1) The \ac{VAE} is trained offline}.
    Lack of alignment in \ac{LSBO} is typically attributed to the purely unsupervised training of \acp{VAE}, which focuses solely on reconstructing the (unlabelled) training data of the \ac{VAE}. In particular, as the highly expressive neural network underlying the \ac{VAE} can distort local neighbourhoods, any smoothness assumptions that motivate Bayesian optimisation in the original space $\mathcal{X}$—for instance, the continuity of a molecule’s performance under small structural perturbations—are unlikely to transfer to smoothness in the latent space of the \ac{VAE}.
    The vast majority of recent \ac{LSBO} work (see Section \ref{sec:related_work}) attempts to fine-tune the \ac{VAE} during optimisation to improve alignment for the current optimisation objective. However, such approaches are plagued by the fundamental challenge of reliably fine-tuning neural networks on small datasets without over-fitting behaviour. Indeed, as the VAE has a large number of hyperparameters from a Bayesian point of view (i.e., the parameters of the decoder), the VAE's decoder is prone to doing all the heavy lifting, which can lead to overfitting analogous to that described for deep kernel learning in \citet{ober2021promises}.
    
\begin{figure}[htp]
    \centering
    \subfigure[\label{fig:pitfalls}]{\includegraphics[width=0.24\textwidth]{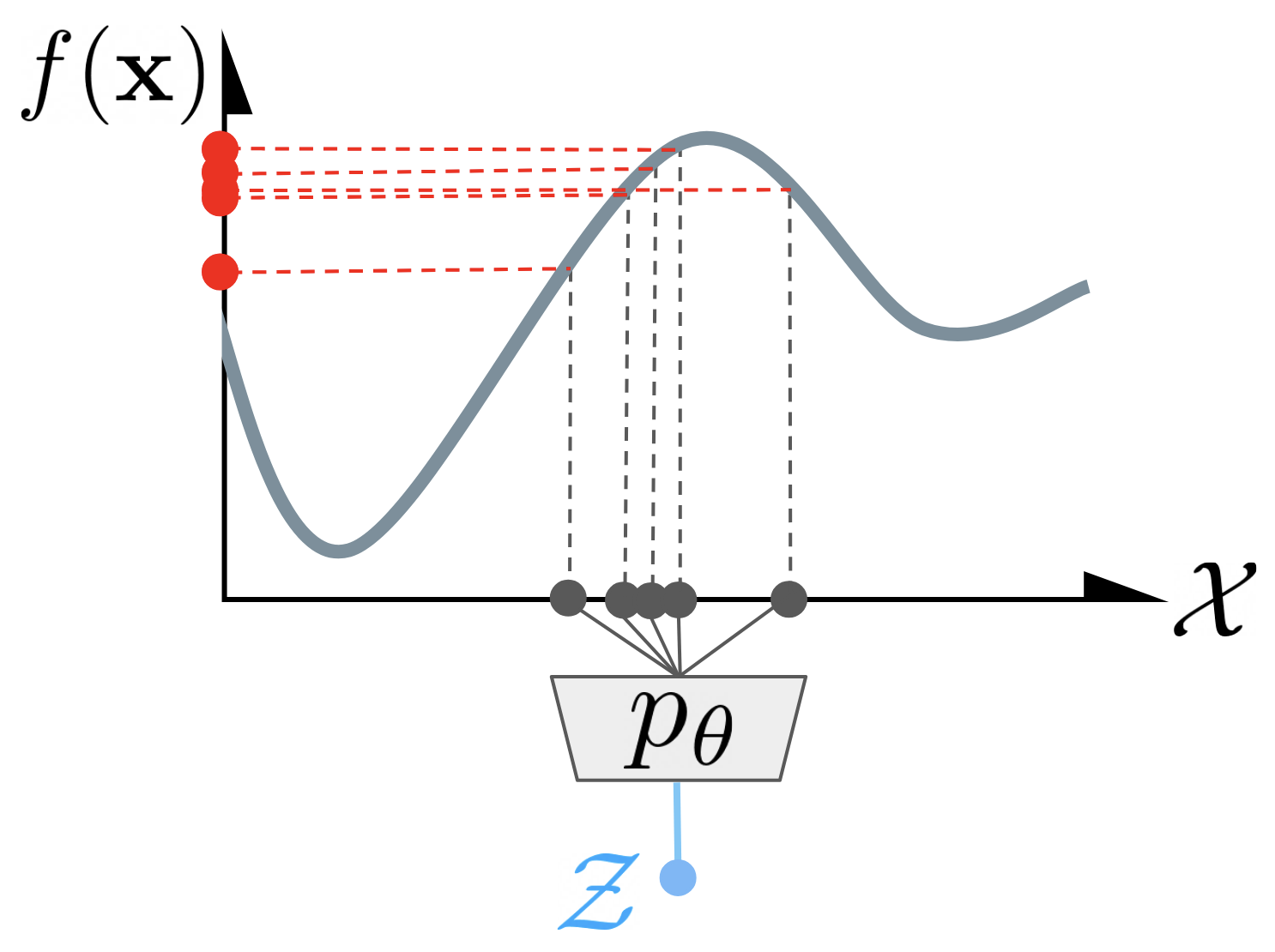}}
    \subfigure[\label{fig:GAT}]{\includegraphics[width=0.19\textwidth]{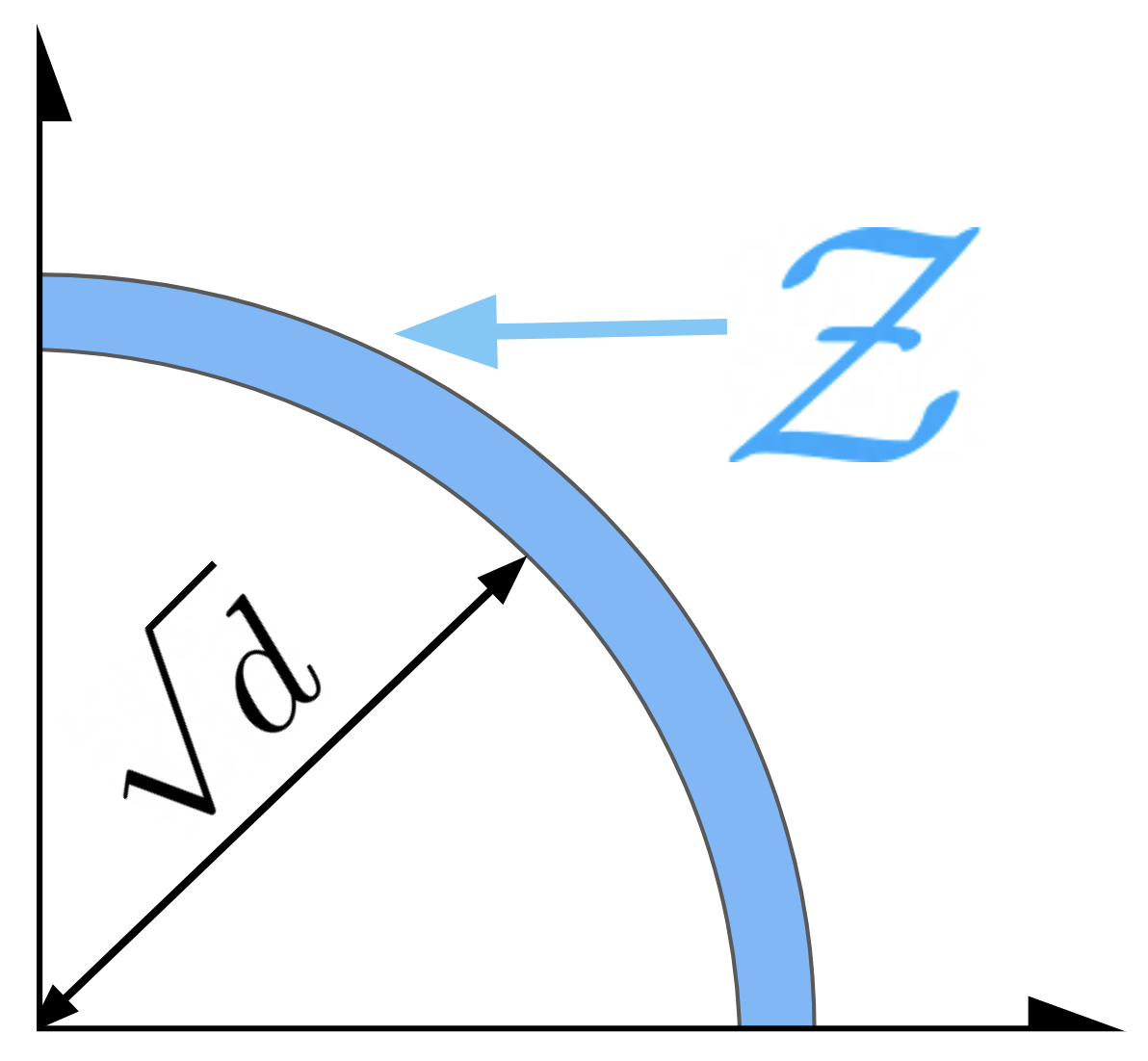}} 
    \caption{(a) In \ac{LSBO}, the same latent input (blue dot) will, via the stochastic decoder (grey box), map to different values in structure space (black dots) and so corresponds to multiple objective function values (red dots) --- a discrepancy that hinders the learning of accurate surrogate models. (b) In higher-dimensional problems, the area of the latent space supported by the prior of the \ac{VAE} (blue) concentrates in a thin circular shell.}
    
\end{figure}

\textbf{2) The \ac{VAE} has a stochastic decoder}. The second challenge for \ac{GP} models in \ac{LSBO}, one largely ignored until the recent works of \citet{chuinversion, leelatent}, is the stochasticity of the decoder. Since the decoder is stochastic, a single latent code $\bm{z}$ can decode to multiple different molecular structures $\bm{x}$, and thereby a range of objective function values (see Figure \ref{fig:pitfalls}). A standard GP can only explain such discrepancies as additional observation noise or by over-fitting, limiting the utility of predictions across the rest of the space --- a well-known problem that has motivated substantial work on  providing GP models that support noisy inputs \citep[e.g.,][]{mchutchon2011gaussian, qing2023robust}.

\textit{In \ac{COWBOYS}, we avoid alignment issues entirely by departing from using a GP in the latent space, instead only fitting surrogate models directly in the data-space.
}

\subsection{LSBO Search Spaces are not Uniform Boxes}
\label{sec:annulus}

The latent spaces of \acp{VAE} are unbounded, therefore \ac{LSBO} typically restricts the search to a pre-specified bounded region $[-\delta,\delta]^d\subset\mathcal{Z}$ (see Algorithm \ref{alg:lsbo}). This simplification serves two main purposes. First, limiting the search space facilitates the use of established acquisition function optimisation methods. Second, by focusing on a central region hypothesized to be more well-behaved, one can limit the presence so-called “dead regions” in the latent space: areas known to yield unrealistic decodings \cite{griffiths2020constrained}. However, for the \acp{VAE} typically considered in molecular design problems, we now show that this intuition is entirely incorrect.

“Dead areas” are regions in a generative model’s latent space that lie far from the training data’s encodings—encodings which themselves populate high-probability zones under the prior. This concept is often referenced when manipulating latent vectors across a wide range of generative models \cite{white2016sampling} (for \acp{VAE} and \acp{GAN}) and \cite{song2020denoising, song2020score} (for diffusion-based approaches). Outside of the \ac{LSBO} setting, common strategies for latent space manipulation draw on a classical result which shows that most of a high-dimensional Gaussian’s probability mass resides in a thin annulus rather than near the centre:

\begin{theorem}[Gaussian Annulus Theorem, Section 3.3.3 of \citet{vershynin2018high}] \label{theorem:GAT} Nearly all of the probability mass of a standardised Gaussian is concentrated in a thin annulus of width $O(1)$ at radius $\sqrt{d}$. \end{theorem}

Although the Gaussian Annulus Theorem is well known in machine learning—leading to “radius-preserving” methods for latent value manipulation like \ac{SLERP} \citep{shoemake1985animating} and sub-space extraction \citep{bodin2024linear}—its implications pose a serious challenge to the practice of clipping the latent search space in \ac{LSBO} when considering \acp{VAE} with even just moderate dimensionality. For example, the encoder of the $128$-dimensional VAE used for our second set of experiments (Section \ref{subsec:hdbo}) maps $95\%$ a $5,000$ random subset of its training data to be within a $128$-dimensional spherical shell of thickness $0.06$. Even advanced \ac{LSBO} approaches, such as adaptive centring of the search space \citep{maus2022local}, still rely on box-shaped regions, making it impossible to target only this high-probability “shell” where the \ac{VAE} encoder produces its most useful structures.

\textit{In COWBOYS we can avoid the need to define a search space entirely, instead proposing a sampling-based strategy rather than one requiring acquisition optimisation.
}

\section{Return of the Latent Space COWBOYS}\label{sec:latentspacecowboys}
As in \ac{LSBO}, COWBOYS relies on two probabilistic models: the \ac{VAE} $\bm{x} \sim p_{\theta}(\bm{x})$ for generating likely valid molecules, and a \ac{GP}-based predictive distribution $p(f(\textbf{x})|D^{\mathcal{X}}_n)$ for estimating objective values. However, rather than predicting these values directly from latent codes—an often challenging task (see above) --- we leverage GP’s proven effectiveness in the original structured space using a dataset $D^{\mathcal{X}}_{n-1}= \{(\bm x_i, y_i)\}_{i=1}^{n-1}$ of structure-evaluation pairs and a Tanimoto kernel (as introduced in Section \ref{sec:bo_intro}).

We now summarise \acfp{COWBOYS}, as Algorithm~\ref{alg:cowboys}. Notably it avoids any explicit reference to the latent space $\mathcal{Z}$. Despite its simplicity, COWBOYS's formulation allows us to exploit the power of the VAE whilst avoiding LSBO's two core challenges: (i) we do not fit models in the latent space and (ii) we do not use the latent space as a search space. To “focus” the generating distribution $p_{\theta}(\bm{x})$ towards promising regions during optimisation, we apply a Bayesian update informed by the GP surrogate. The algorithm proceeds via two main steps, described below for the $n^{th}$ iteration under an initial design of size $N_{\textrm{init}}$.

\begin{enumerate}
    \item \textbf{Initial design} ($n\leq N_{\textrm{init}}$) To generate our initial design of size $N_{\textrm{init}}$ we use the VAE exactly as it was designed, i.e. by decoding Gaussian latent samples. Therefore for $n\in\{1, ..., N_{\textrm{init}}\}$ we sample
\begin{equation}
    \bm x_n \sim p_{\theta}(\bm x), \label{eq:init_goal}
\end{equation}
resulting in an initial dataset of structure-evaluation pairs over which we can initialise our GP model, without the need to specify a clipped search space.

\item 
\textbf{Optimisation steps} ($n> N_{\textrm{init}}$) After the initial design, \textsc{COWBOYS} refines its search by only sampling molecules that the surrogate GP predicts will exceed the best observed objective value so far. More concretely, 
\begin{equation}
 \bm x_n\sim p_{\theta}(\bm x|f_{\bm x}>f^*, D_n^{\mathcal{X}}), \label{eq:opt_goal} 
\end{equation} 
where $f_{\bm x} | D_n^{\mathcal{X}}$ is a random variable following the GP’s posterior predictive distribution for the objective value at \(\mathbf{x}\), and \(f^*\) denotes the highest observed objective value so far. 
\end{enumerate}

 To build intuition for \textsc{COWBOYS}, one might consider a simple, though computationally prohibitive, approach via rejection sampling: generate a large number of candidate structures from the VAE and retain only those satisfying $f_{\bm x}>f^*| D_n^{\mathcal{X}}$ (under a realisation of $f_{\bm x}| D_n^{\mathcal{X}}$). 
 
 \textbf{Batch BO}. Note that an arbitrary number of samples can be drawn in parallel from (\ref{eq:opt_goal}), making COWBOYS well-suited for the large-batch evaluations often tackled in Bayesian optimisation \citep{vakili2021scalable} and active learning \citep{ober2025big}.


\begin{algorithm}[tb]
   \caption{\label{alg:cowboys}COWBOYS}
\begin{algorithmic}
   \STATE {\bfseries Input:} Budget $N$, init size $N_{\textrm{init}}$
   \FOR{$n\in\{1,..,N\}$}
   \IF[initial design]{$n<N_{\textrm{init}}$}
   \STATE $\bm x_n \sim p_{\theta}(\bm x)$ \COMMENT{vanilla VAE}
   \ELSE[sequential optimisation]
   \STATE $f^* \leftarrow \max_{i=1,..,n-1}y_i$
   \STATE $\bm x_n \sim p_{\theta}(\bm x|f_{\bm x}>f^*, D_{n-1}^{\mathcal{X}})$\hspace{-2cm} \COMMENT{GP-conditioned
   VAE}
   \ENDIF
   \STATE Evaluate new structure $y_n\leftarrow f(\bm{x}_n)$
    \STATE Update dataset $D^{\mathcal{X}}_n\leftarrow D^{\mathcal{X}}_{n-1} \bigcup \{(\bm x_n,y_n)\}$
    \STATE Fit structured space GP on $D^{\mathcal{X}}_n$
   \ENDFOR
    \STATE \textbf{return} Believed optimum across $\{\bm{x}_1,\dots,\bm{x}_n\}$
\end{algorithmic}
\end{algorithm}

\section{Practical Sampling for COWBOYS}
\label{sec:practical}

Although the GP-conditioned VAE is a conceptually appealing way to focus on increasingly narrower, relevant proportions of molecular space, sampling from (\ref{eq:opt_goal}) remains non-trivial. Indeed, the naive rejection sampler that uses the VAE $p_{\theta}(\mathbf{x})$ as a proposal becomes prohibitively inefficient once the region of the search space likely to satisfy $f_{\textbf{x}}>f^*_n$ narrows, which typically occurs in the early stages of the optimisation process. Consequently, we recommend leveraging more advanced sampling algorithms from the computational statistics literature.

In the remainder of this section, we illustrate how a popular Markov Chain Monte Carlo \citep[MCMC]{brooks2011handbook} approach can be used to approximate COWBOYS. Specifically, we demonstrate that sampling from COWBOYS’ GP-conditioned \acf{VAE} can be reformulated as sampling from a posterior distribution induced by a Gaussian prior and a corresponding likelihood --- a well-established setting for which effective MCMC algorithms already exist and can be readily applied.

\subsection{COWBOYS $\approx$ Inference under a Gaussian Prior}

Our practical implementation of COWBOYS approximates the GP-conditioned VAE \eqref{eq:opt_goal} as the posterior distributions induced by a specific likelihood and a (potentially high-dimensional) Gaussian prior. However, this approximation strategy requires a deterministic mapping from the latent space to the structure space, rather than the stochastic mapping provided by the VAE’s decoder. Consequently, COWBOYS only ever considers the most likely decoding of the latent variable, inducing a deterministic mapping between latent codes and structure given by $h_{\theta}(\bm z) = \argmax_{\bm x}p_{\theta}(\bm x | \bm z)$. In other words, we approximate the VAE’s decoder distribution with the delta function:
\begin{align}
    \hat{p}_{\theta}(\bm x | \bm z) \approx \delta(\bm x - h_{\theta}(\bm z)), \nonumber
\end{align}
i.e. a deterministic decoding strategy that selects only the most-likely decoded molecule from each latent location \textbf{z}.

Note that we are not the first to replace the VAE’s probabilistic decoding with the most-likely mapping (e.g., \citet{gonzalez2024survey}). Removing this source of variation can, at the risk of reducing the diversity of candidate samples, mitigate the alignment issues discussed in Section \ref{section:poor_model}. In our experiments, however, we found that COWBOYS did not suffer empirically from this reduced variation (see Appendix \ref{appendix:ablate}). We also stress that the VAE is still initially trained with a stochastic decoder, and that it is only when performing BO that we take the most likely mapping.

Once we have defined the mapping $h_{\theta}:\mathcal{Z}\rightarrow\mathcal{X}$, each latent variable $\bm z\in\mathcal{Z}$ now defines a corresponding structure $\bm x\in\mathcal{X}$. Consequently, rather than sampling a new structure directly, we can reframe the sampling task in \eqref{eq:opt_goal} as sampling a new latent value $\bm z$ from an appropriate distribution and then mapping $\bm z$ through $h_{\theta}$ to obtain the new structure. Specifically, under the deterministic approximation of the decoder distribution, sampling $\bm x $ from \eqref{eq:opt_goal} is equivalent to the following two step procedure:
\framebox{
\begin{minipage}{22em}
\begin{align}
    \bm z& \sim p(\bm z | g_{\theta, \bm z} > f^*,D_n^{\mathcal{X}})\label{eq:cowboys_z}\\
    \bm x& = h_{\theta}(\bm z), \nonumber
\end{align}
\end{minipage} 
}
where $g_{\theta, \bm z} | D_n^{\mathcal{X}}$ is the random variable following the GP posterior predictive distribution for the objective value at the decoded structure \(\bm{x}=h_{\theta}(\bm z)\), i.e., $g_{\theta, \bm z} = f(h_{\theta}(\bm z))$.

By decomposing \eqref{eq:cowboys_z} via Bayes' rule as
\begin{align} p(\bm z | g_{\theta,\bm z} > f^*,D_n^{\mathcal{X}}) \propto p\bigl(g_{\theta,\bm z} > f^* \mid D^{\mathcal{X}}_n\bigr)\times p(\bm z), \label{eq:decomp}
\end{align} we can now see that \eqref{eq:cowboys_z} corresponds to sampling from the posterior distribution induced by the VAE's Gaussian prior $p(\bm z)$ and a likelihood  $p\bigl(g_{\theta,\bm z} > f^* \mid D^{\mathcal{X}}_n\bigr)$, which resembles the PI acquisition function (and so can be calculated easily in closed-form under the GP). 
Hence, our approximation of COWBOYS reduces to a classical form for which many well-established sampling methods are available.

\subsection{Preconditioned Crank-Nicolson MCMC}
For our experiments, we sample from \eqref{eq:cowboys_z} using a modification of the  \ac{PCN} algorithm \citep{cotter2013mcmc}. PCN is well-suited to sampling from medium- to high-dimensional Gaussian priors, as it concentrates on the annulus of high-prior probability (as discussed in Section \ref{sec:annulus}) rather than attempting to explore the entire parameter space. This contrasts with more conventional \ac{MCMC} algorithms (e.g., random-walk methods), which rapidly deteriorate in effectiveness as dimensionality increases \citep{hairer2014spectral} --- an observation closely aligned with our critique of existing \ac{LSBO} strategies.

Exact details of our implementation of \ac{PCN} are provided in Appendix \ref{appendix:sampling}. Note that, in order to leverage parallel hardware, we introduced a minor alteration that transforms PCN into a non-reversible MCMC algorithm, thereby forfeiting the rigorous sample quality guarantees derived by the computational statistics community. However, in the context of COWBOYS, where our ultimate goal is Bayesian optimisation, our empirical findings suggest that producing a set of approximate samples is sufficient.

\subsection{Computational Complexity}
We now analyse the computational overhead incurred by COWBOYS and standard LSBO in non-batch setting. Ignoring the shared computational cost of fitting GP surrogate models, the $n^{th}$ step of a COWBOYS algorithm using PCN with $C$ MCMC chains and $S$ MCMC steps has complexity $O(CS(N^2 + V))$, corresponding to the need to make a GP prediction $O(N^2)$ (assuming a cached Gram matrix inverse) and VAE decoding ( at cost $V$) for each point considered by PCN. In contrast, standard LSBO incurs $O(AN^2 + V)$, where $A$ are the number of evaluations required to maximise its acquisition function. We set up our experiments via choices of $C$ and $S$ so that GP costs are equivalent ($C\times S\approx A$). We stress that although COWBOYS requires more VAE evaluations than standard LSBO, advanced LSBO approaches that fine-tune the VAE on incoming data also require large numbers of additional VAE evaluations.





\section{Baselines \& Related Works}
\label{sec:related_work}

\textbf{Improving ``alignment"}. The primary recent focus of the LSBO literature has been on allowing VAEs to be fine-tuned during optimisation, encouraging the alignment of latent spaces with respect to the optimisation objective, with \citet{eissman2018bayesian}, \citet{tripp2020sample}, \citet{moriconi2020high}, \citet{grosnit2021high}, \citet{maus2022local}, and \citet{chen2024pg} proposing methods for fine-tuning the latent space mappings --- using additional supervised neural networks, re-weighted losses, or performing joint inference over the surrogate model/mapping. Most recently, \citet{lee2023advancing} fine-tune VAEs with the explicit goal of ensuring the optimisation objective is smooth (in a Lipschitz sense) across the latent space --- an approach extended further in \citet{chuinversion} to also address the additional alignment issue arising from stochasticity in the VAE's decoder. Here, LSBO is recast as an inverse problem, where they use optimisation identify the latent codes most likely to decode to any suggested structures before fitting surrogate models. 

\textbf{Encouraging valid decodings}. Custom VAE decoders have been proposed that can ensure the satisfaction of string-based \citep{kusner2017grammar} or graph-based constraints \citep{jin2018junction}.  Alternatively, the suggestion of valid structures, can be encouraged by using an additional model that predicts the validity of proposed structures to help triage the query points suggested by BO \citep{griffiths2020constrained} or by avoiding areas of the latent space where the decoder's epistemic uncertainty is high \citep{notin2021improving}.

\textbf{Additional improvements for LSBO}. Extensions support high-dimensional latent spaces \citep{maus2022local}, multi-objective  \citep{stanton2022accelerating} and batch optimisation \citep{maus2022discovering}, as well as a method to include structure-level knowledge into the surrogate model alongside latent representations \citep{deshwal2021combining}. Recent work of \citep{ramchandranhigh} proposed the use of GP VAEs that allows the inclusion of auxiliary variables to help learn useful latent spaces.





\section{Results}

\begin{figure}
    \centering
    \subfigure[Osimertinib MPO]{\includegraphics[height=0.16\textwidth]{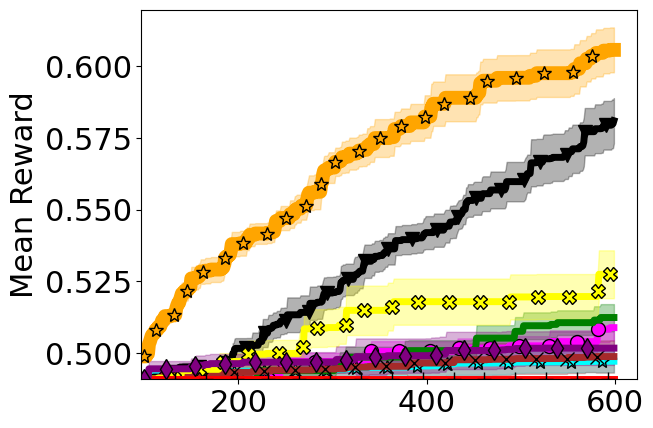}}%
    \subfigure[Median Molecules 2]{\includegraphics[height=0.16\textwidth]{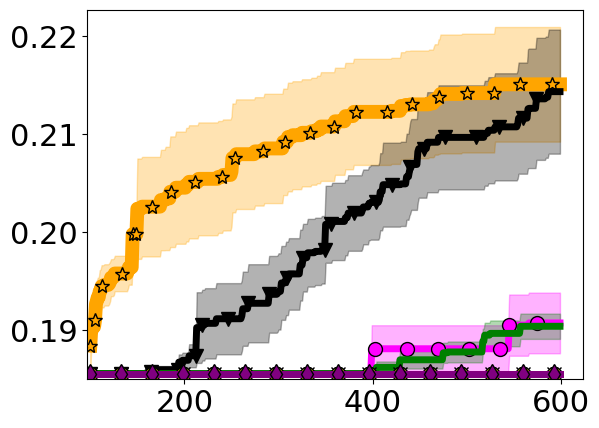}}
    \subfigure[Amlodipine MPO]{\includegraphics[height=0.16\textwidth]{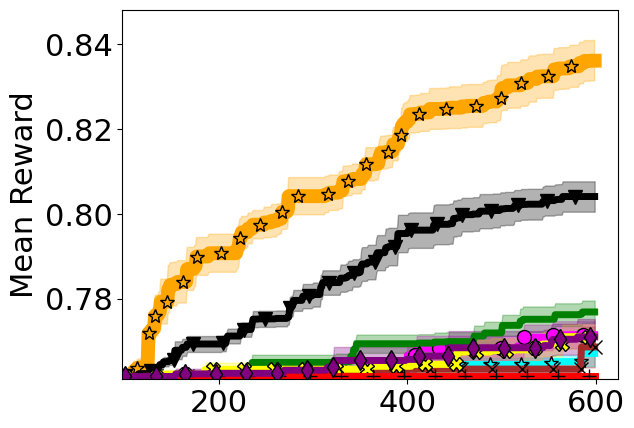}}%
    \subfigure[Perindopril MPO]{\includegraphics[height=0.16\textwidth]{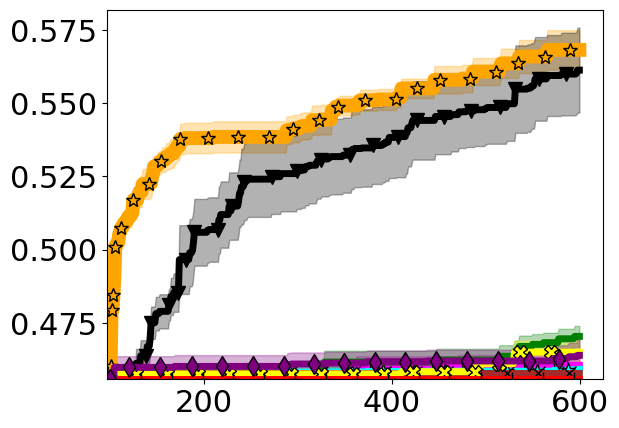}}
    \subfigure[Ranolazine MPO]{\includegraphics[height=0.175\textwidth]{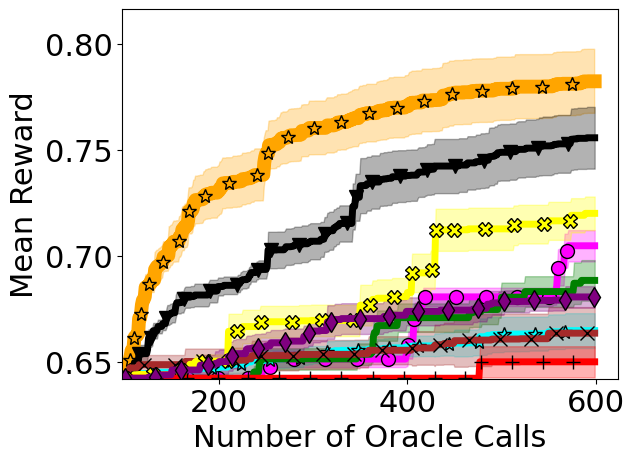}}%
    \subfigure[Zaleplon MPO]{\includegraphics[height=0.175\textwidth]{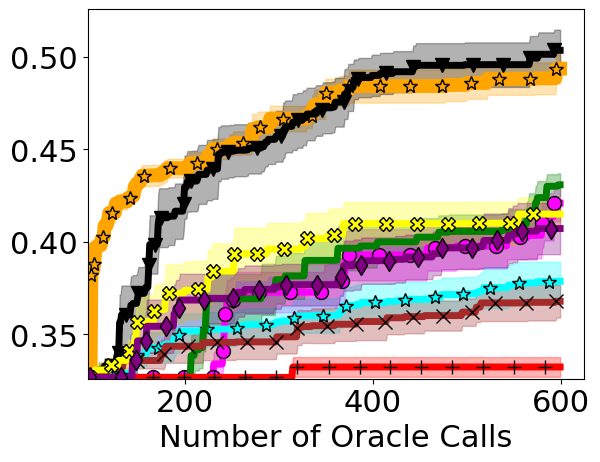}}
   \includegraphics[width=0.48\textwidth]{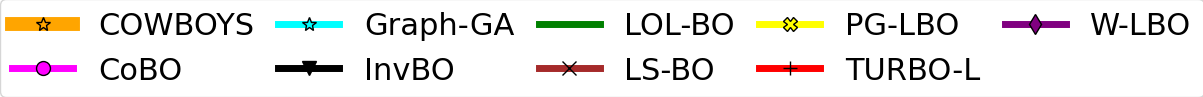}
\caption{Average performance ($\pm$ standard error) of COWBOYS over 10 runs on problems considered by \cite{chuinversion}.  COWBOYS achieves a substantial improvement in sample efficiency over all existing LSBO methods (including those able to fine-tune VAEs to incoming data) in this low data regime, with only the recently proposed InvBO sometimes matching its performance. We stress that COWBOYS does not fine-tune its VAE during optimisation, rather just uses it more efficiently.}%
\label{fig::results_adaptive}%
\end{figure}

 We compare the performance of COWBOYS over multiple VAES, against 24 unique algorithms, and on 16 open-source benchmark problems for molecular design. Our experiments replicate the exact experimental setups of \citet{chuinversion}, \citet{gonzalez2024survey}, and \citet{maus2024approximation}; results are shown in Figure~\ref{fig::results_adaptive}, Table~\ref{table:hdbo}, and Figure~\ref{fig::results_fixed}, respectively. See \cite{brown2019guacamol} for the scientific motivation and clear practical details behind these benchmarks.

\textbf{General details.} Our code is integrated into the benchmarking suite of \citet{gonzalez2024survey} and is available at \emph{https://github.com/henrymoss/ROTLSC}. We build on \texttt{BoTorch} \citep{balandat2020botorch}, \texttt{GPyTorch} \citep{gardner2018gpytorch}, and lastly \texttt{GAUCHE} \citep{griffiths2024gauche} for its Tanimoto and molecule utilities, adopting all default settings for model optimisation and kernel hyperparameters. Baseline methods are drawn from their respective open-source implementations. For \ac{COWBOYS}'s MCMC-based sampling strategy (our modified PCN, described in Appendix~\ref{appendix:sampling}), we run 100 MCMC steps over a single chain when the batch size is one ($B=1$), and 50 chains when $B>5$. This matches the $5,000$ acquisition function evaluation budget typically used by TURBO \citep{eriksson2019scalable}, a key component of all other high-performance LSBO routines. Appendix \ref{appendix:ablate} examines COWBOYS' robustness to these MCMC settings and highlights the combined impact of its two primary components by applying COWBOYS-style sampling but with a latent-space GP model.

\subsection{Comparison with State of the Art LSBO Methods}
\label{sec:lsbobest}

\textbf{Baselines}. Following \citet{chuinversion}, we consider \textbf{LOL-BO} \citep{maus2022local}, \textbf{CoBO} \citep{lee2024advancing}, \textbf{W-LBO} \citep{tripp2020sample}, \textbf{InvBO} \citep{chuinversion}, and \textbf{PG-LBO} \citep{chen2024pg}, all of which require fine-tuning the VAE during optimisation. We also evaluate the standard \textbf{LSBO} method (Algorithm \ref{alg:lsbo}) and its trust region variant \textbf{TURBO-L} \citep{eriksson2019scalable}. As a reference point, we include \textbf{Graph-GA} \citep{jensen2019graph}, a BO-free baseline that uses a widely adopted graph-based genetic algorithm. For a more detailed discussion of these baselines, see \citet{chuinversion}. Note that we used a pre-trained VAE provided via direct correspondence from the authors of \cite{maus2022local}, as the one provided in their open-source implementation was not sufficient to recreate their results.

\textbf{Results}. As shown in Figure~\ref{fig::results_adaptive}, COWBOYS achieves marked improvements in sample efficiency over existing LSBO methods in the low-data regime, outperforming even those permitted to fine-tune their VAE. We replicate the experimental setup from \citet{maus2022local,maus2022discovering,lee2024advancing,chuinversion}, focusing on lower-budget problems (COWBOYS cannot currently handle evaluation budgets of 80,000 due to a current lack of availability of scalable GP models for discrete data; see Section~\ref{sec:discussion}). Across six tasks introduced by \citet{brown2019guacamol}, we run 100 BO steps with a batch size $B=5$, starting from 100 representative molecules as defined by \citet{lee2024advancing}, a scenario designed to mimic real-world lead optimisation. These experiments employ the SELFIES VAE of \citet{maus2022local}, which encodes and decodes molecules into a 256-dimensional latent space using six transformer layers \citep[see][for details]{maus2022local}.


\subsection{High-dimensional BO Benchmark}
\label{subsec:hdbo}

\textbf{Baselines}. Following the setup in \citet{gonzalez2024survey}, we compare COWBOYS to three categories of methods: (i) non-BO evolutionary optimisers applied to the VAE latent space: \textbf{hill-climbing}, \textbf{CMAES}, and a Genetic Algorithm (\textbf{GA}), (ii) high-dimensional BO methods also operating in the VAE latent space: \textbf{RandomLineBO} \citep{kirschner2019adaptive} and \textbf{TURBO} \citep{eriksson2019scalable}, and (iii) BO methods that optimise directly in the discrete (molecular) space: \textbf{Bounce} \citep{papenmeier2023bounce} and \textbf{ProbRep} \citep{daulton2022bayesian}. For a more detailed discussion of these baselines, see the online documentation in \citet{gonzalez2024survey}.

\textbf{Results}. Table~\ref{table:hdbo} shows that COWBOYS consistently outperforms all compared methods on the high-dimensional discrete sequence optimisation tasks provided by \citet{gonzalez2024survey}, (based on the PMO benchmark of \citet{gao2022sample}). All algorithms start from an initial design of 10 molecules and run for 300 BO steps with a batch size of $B=1$. Because the original VAE used by \citet{gonzalez2024survey} had relatively poor reconstruction accuracy, we reran the entire benchmark (and all baselines) with an updated, more expressive, 128-dimensional fully-connected VAE using the latest release of the benchmarking suite.

\begin{table*}
\tabcolsep=0.1cm
\scalebox{0.85}{
\begin{tabular}{l|l|lll|ll|ll}
\hline
Objective function                   & \texttt{COWBOYS}                            & \texttt{HillClimbing}                       & \texttt{CMAES}                               & \texttt{GA}                                 & \texttt{RandomLineBO}                        & \texttt{TURBO}                              & \texttt{Bounce}                              & \texttt{ProbRep}                             \\ \hline
albuterol\_similarity     & \cellcolor[HTML]{9AFF99}0.472±0.08 & \cellcolor[HTML]{009901}0.487±0.06 & \cellcolor[HTML]{9AFF99}0.453±0.04  & 0.356±0.05                         & \cellcolor[HTML]{9AFF99}0.454±0.06  & \cellcolor[HTML]{9AFF99}0.456±0.04 & 0.16 ± 0.01                         & 0.21 ± 0.03                         \\
amlodipine\_mpo           & \cellcolor[HTML]{009901}0.477±0.04 & 0.449±0.02                         & \cellcolor[HTML]{9AFF99}0.458±0.02  & 0.440±0.01                         & 0.453±0.02                          & 0.444±0.02                         & 0.00 ± 0.00                         & 0.00 ± 0.00                         \\
celecoxib\_rediscovery    & \cellcolor[HTML]{009901}0.217±0.02 & 0.202±0.01                         & \cellcolor[HTML]{9AFF99}0.213±0.02  & 0.202±0.01                         & 0.204±0.01                          & 0.202±0.01                         & 0.02 ± 0.01                         & 0.02 ± 0.00                         \\
deco\_hop                 & \cellcolor[HTML]{009901}0.570±0.01 & \cellcolor[HTML]{9AFF99}0.562±0.01 & 0.563±0.00                          & \cellcolor[HTML]{9AFF99}0.562±0.01 & \cellcolor[HTML]{9AFF99}0.564±0.01  & \cellcolor[HTML]{9AFF99}0.562±0.01 & 0.50 ± 0.00                         & 0.51 ± 0.00                         \\
drd2\_docking             & \cellcolor[HTML]{9AFF99}0.342±0.17 & 0.097±0.09                         & 0.087±0.09                          & 0.076±0.10                         & \cellcolor[HTML]{009901}0.346±0.37  & 0.170±0.10                         & 0.01 ± 0.00                         & 0.01 ± 0.00                         \\
fexofenadine\_mpo         & \cellcolor[HTML]{009901}0.682±0.02 & \cellcolor[HTML]{9AFF99}0.644±0.04 & \cellcolor[HTML]{9AFF99}0.652±0.05  & 0.578±0.06                         & \cellcolor[HTML]{9AFF99}0.669±0.03  & \cellcolor[HTML]{9AFF99}0.632±0.06 & 0.13 ± 0.13                         & 0.20 ± 0.08                         \\
gsk3\_beta                & \cellcolor[HTML]{009901}0.368±0.03 & 0.252±0.05                         & 0.321±0.11                          & 0.241±0.04                         & 0.170±0.05                          & 0.302±0.05                         & 0.09 ± 0.08                         & 0.12 ± 0.02                         \\
isomer\_c7h8n2o2          & \cellcolor[HTML]{009901}1.000±0.00 & 0.788±0.19                         & 0.872±0.07                          & 0.812±0.10                         & \cellcolor[HTML]{9AFF99}0.864±0.15  & 0.880±0.07                         & 0.11 ± 0.09                         & 0.24 ± 0.11                         \\
isomer\_c9h10n2o2pf2cl    & \cellcolor[HTML]{009901}0.690±0.08 & 0.600±0.12                         & 0.567±0.11                          & \cellcolor[HTML]{9AFF99}0.639±0.07 & 0.654±0.12                          & 0.618±0.24                         & 0.01 ± 0.01                         & 0.06 ± 0.03                         \\
jnk3                      & \cellcolor[HTML]{009901}0.138±0.05 & 0.122±0.04                         & \cellcolor[HTML]{9AFF99}0.136±0.02  & 0.130±0.04                         & 0.126±0.04                          & 0.092±0.03                         & 0.05 ± 0.04                         & 0.06 ± 0.01                         \\
median\_1                 & \cellcolor[HTML]{009901}0.198±0.02 & 0.162±0.01                         & 0.187±0.03                          & 0.137±0.02                         & 0.174±0.03                          & 0.154±0.04                         & 0.03 ± 0.01                         & 0.02 ± 0.00                         \\
median\_2                 & \cellcolor[HTML]{009901}0.162±0.01 & 0.146±0.01                         & 0.149±0.01                          & 0.145±0.01                         & 0.145±0.01                          & 0.145±0.01                         & 0.01 ± 0.00                         & 0.01 ± 0.00                         \\
mestranol\_similarity     & \cellcolor[HTML]{009901}0.384±0.01 & 0.348±0.02                         & 0.362±0.02                          & 0.266±0.03                         & 0.353±0.01                          & \cellcolor[HTML]{9AFF99}0.378±0.07 & 0.01 ± 0.00                         & 0.02 ± 0.00                         \\
osimetrinib\_mpo          & \cellcolor[HTML]{009901}0.722±0.03 & \cellcolor[HTML]{9AFF99}0.718±0.02 & \cellcolor[HTML]{9AFF99}0.711±0.05  & \cellcolor[HTML]{9AFF99}0.707±0.04 & \cellcolor[HTML]{9AFF99}0.714±0.03  & \cellcolor[HTML]{9AFF99}0.711±0.03 & 0.30 ± 0.31                         & 0.59 ± 0.04                         \\
perindopril\_mpo          & \cellcolor[HTML]{009901}0.382±0.04 & \cellcolor[HTML]{9AFF99}0.326±0.11 & \cellcolor[HTML]{9AFF99}0.335±0.11  & \cellcolor[HTML]{9AFF99}0.334±0.12 & \cellcolor[HTML]{9AFF99}0.337±0.11  & \cellcolor[HTML]{9AFF99}0.362±0.06 & 0.00 ± 0.00                         & 0.00 ± 0.00                         \\
ranolazine\_mpo           & \cellcolor[HTML]{009901}0.648±0.04 & 0.609±0.02                         & \cellcolor[HTML]{9AFF99}0.625±0.08  & 0.484±0.06                         & \cellcolor[HTML]{9AFF99}0.632±0.03  & 0.597±0.04                         & 0.00 ± 0.00                         & 0.11 ± 0.02                         \\
rdkit\_logp               & 10.678±0.99                        & 11.383±6.12                        & \cellcolor[HTML]{9AFF99}14.474±7.41 & 9.307±1.67                         & \cellcolor[HTML]{009901}19.484±5.92 & 9.691±0.71                         & 3.12 ± 1.20                         & 5.49 ± 3.01                         \\
rdkit\_qed                & 0.909±0.03                         & 0.902±0.03                         & \cellcolor[HTML]{009901}0.912±0.03  & 0.902±0.03                         & 0.906±0.04                          & 0.903±0.04                         & 0.52 ± 0.09                         & \cellcolor[HTML]{FFFFFF}0.60 ± 0.05 \\
sa\_tdc                   & 7.909±0.07                         & 7.920±0.06                         & 7.516±0.55                          & \cellcolor[HTML]{009901}8.878±0.16 & 7.877±0.11                          & 7.952±0.05                         & \cellcolor[HTML]{9AFF99}8.36 ± 0.46 & \cellcolor[HTML]{FFFFFF}8.59 ± 0.13 \\
scaffold\_hop             & \cellcolor[HTML]{009901}0.435±0.01 & \cellcolor[HTML]{9AFF99}0.427±0.01 & \cellcolor[HTML]{9AFF99}0.430±0.01  & \cellcolor[HTML]{9AFF99}0.427±0.01 & \cellcolor[HTML]{9AFF99}0.427±0.01  & \cellcolor[HTML]{9AFF99}0.429±0.01 & 0.34 ± 0.01                         & 0.34 ± 0.00                         \\
sitagliptin\_mpo          & \cellcolor[HTML]{009901}0.259±0.07 & 0.151±0.13                         & \cellcolor[HTML]{9AFF99}0.237±0.12  & 0.154±0.09                         & \cellcolor[HTML]{9AFF99}0.233±0.12  & \cellcolor[HTML]{9AFF99}0.163±0.12 & 0.00 ± 0.00                         & 0.00 ± 0.00                         \\
thiothixene\_rediscovery  & \cellcolor[HTML]{009901}0.247±0.03 & 0.224±0.02                         & 0.224±0.02                          & 0.224±0.02                         & 0.226±0.02                          & 0.224±0.02                         & 0.02 ± 0.01                         & 0.03 ± 0.01                         \\
troglitazone\_rediscovery & \cellcolor[HTML]{009901}0.205±0.02 & \cellcolor[HTML]{9AFF99}0.187±0.02 & \cellcolor[HTML]{9AFF99}0.203±0.03  & \cellcolor[HTML]{9AFF99}0.187±0.02 & \cellcolor[HTML]{9AFF99}0.188±0.02  & \cellcolor[HTML]{9AFF99}0.195±0.01 & 0.02 ± 0.01                         & 0.02 ± 0.00                         \\
valsartan\_smarts         & \cellcolor[HTML]{009901}0.000±0.00 & \cellcolor[HTML]{009901}0.000±0.00 & \cellcolor[HTML]{009901}0.000±0.00  & \cellcolor[HTML]{009901}0.000±0.00 & \cellcolor[HTML]{009901}0.000±0.00  & \cellcolor[HTML]{009901}0.000±0.00 & \cellcolor[HTML]{009901}0.00 ± 0.00 & \cellcolor[HTML]{009901}0.00 ± 0.00 \\
zaleplon\_mpo             & \cellcolor[HTML]{009901}0.379±0.05 & \cellcolor[HTML]{9AFF99}0.337±0.08 & \cellcolor[HTML]{9AFF99}0.293±0.12  & \cellcolor[HTML]{9AFF99}0.297±0.11 & \cellcolor[HTML]{9AFF99}0.344±0.08  & \cellcolor[HTML]{9AFF99}0.297±0.11 & \cellcolor[HTML]{FFFFFF}0.00 ± 0.00 & \cellcolor[HTML]{FFFFFF}0.00 ± 0.00 \\ \hline
\end{tabular}}
\caption{Average performance ($\pm$ s.d.) over 5 repetitions of COWBOYS on the discrete BO benchmarking suite of \cite{gonzalez2024survey}. We stress best average scores achieved after 300 evaluations (dark) and scores within a single standard deviation of best (light).}
\label{table:hdbo}
\end{table*}

\subsection{Comparison with Traditional LSBO}

\textbf{Baselines}. In our final set of experiments, we restrict our focus to methods that do not fine-tune their VAE. Specifically, we run COWBOYS and LSBO/TURBO-L with exact Gaussian Processes for 100 steps of $B=20$ evaluations, a limit imposed by the computational demands of exact GP inference. We also include scalable variants—\textbf{EULBO EI} and \textbf{EULBO KG} \citep{maus2024approximation}, \textbf{Vanilla BO} \citep{hvarfner2024vanilla}, and \textbf{IPA} \citep{moss2022information, moss2023inducing}—that rely on Sparse Variational Gaussian Processes (SVGP) \citep{hensman2013gaussian} to extend LSBO to 4,000 steps. This allows us to examine whether significantly larger optimisation budgets enable these methods to match COWBOYS’ performance. Extending COWBOYS to accommodate large-budget optimisation is an intriguing avenue for future work (Section \ref{sec:discussion}). Unlike above, here we used the exact pre-trained VAE provided the open-source implementation of \citep{maus2022local} for all approaches.

\begin{figure}
    \centering
    \subfigure[Osimertinib MPO]{\includegraphics[height=0.16\textwidth]{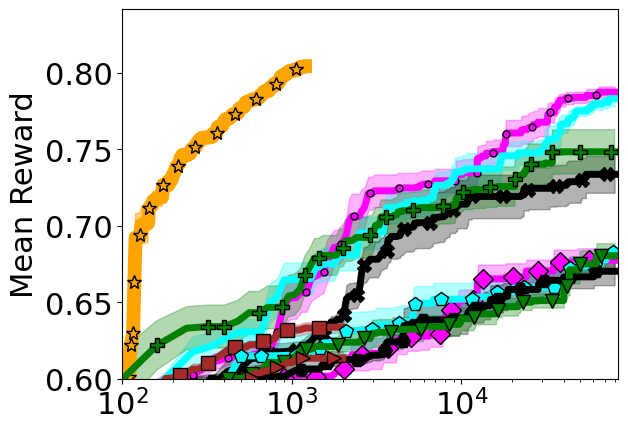}} 
    \subfigure[Fexofenadine MPO]{\includegraphics[height=0.16\textwidth]{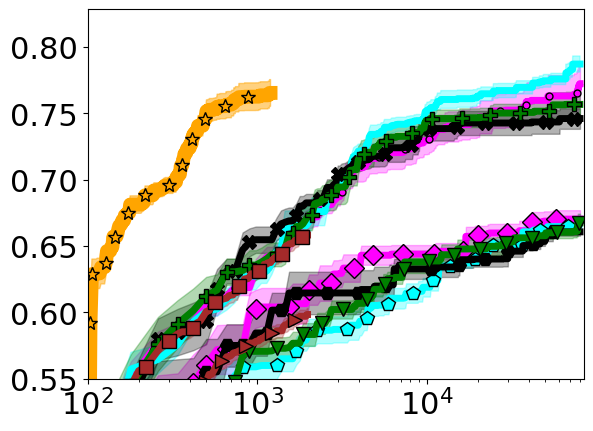}} 
    \subfigure[Median Molecules 1]{\includegraphics[height=0.173\textwidth]{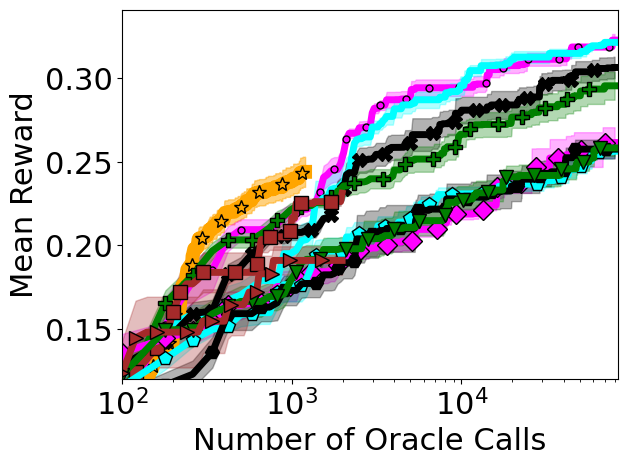}} 
    \subfigure[Median Molecules 2]{\includegraphics[height=0.173\textwidth]{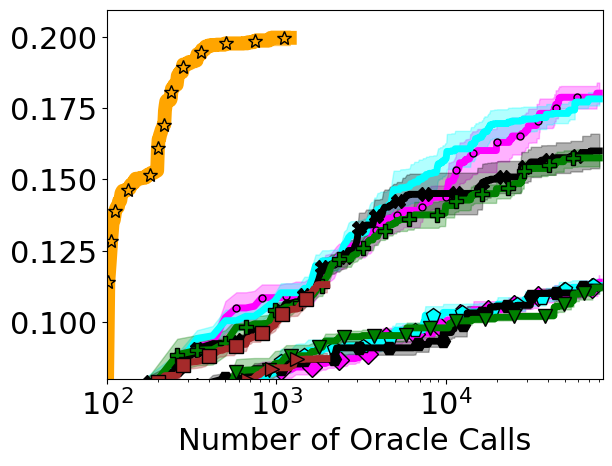}} 
    \includegraphics[width = 0.45\textwidth]{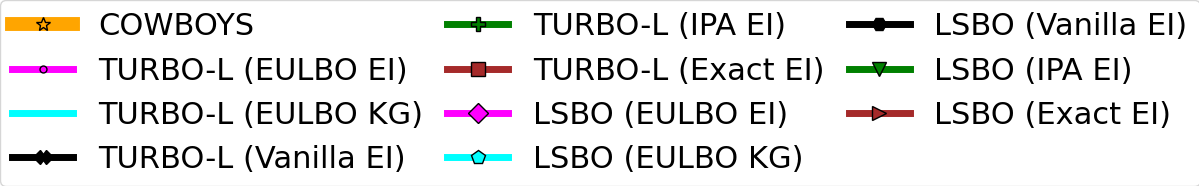}
\caption{Average performance ($\pm$ standard error) over 20 repetitions with an log-scaled x-axis, demonstrating that, among LSBO methods that cannot fine-tune their latent space, COWBOYS provides significant improvemnt in efficiency. }%
\label{fig::results_fixed}%
\end{figure}

\textbf{Results}. As shown in Figure~\ref{fig::results_fixed}, COWBOYS achieves substantially better performance than other LSBO methods that also do not fine-tune the VAE. Even when given a two orders-of-magnitude increase in optimisation budget, these baselines do not catch up to COWBOYS, highlighting the value of our proposed method of combining VAEs and GPs for  Bayesian optimisation. These experiments replicate those in \citet{maus2024approximation}, who perform LSBO using the latent space of a SELFIES VAE (see Section~\ref{sec:lsbobest}), focusing on a subset of the molecular design benchmarks discussed earlier.

\section{Discussion and Limitations}
\label{sec:discussion}
COWBOYS leverages a simple Bayesian updating mechanism that links separately trained VAEs and GPs through a sampling strategy progressively biased toward promising structures. Despite its simplicity, this sampling-based approach directly addresses LSBO's two primary challenges. First, COWBOYS \emph{does not} model the objective in the VAE latent space; instead, it retains the GP in the original, structured representation of the data, allowing the use of specialized kernels. Second, COWBOYS \emph{does not} employ the latent space for optimisation; instead, it naturally restricts its focus to regions that are both plausible under the VAE prior and potentially high-performing under the GP. We demonstrate that COWBOYS offers a substantial boost in sample efficiency on low-budget molecular design benchmarks.

\textbf{Extensions into new application areas}. By working in the raw structure space, our approach naturally supports specialized kernels tailored to particular domains—an area where there is a rich, yet underused, literature. For instance, the Tanimoto kernel, originally proposed by chemists, encodes prior knowledge of what sort of molecular attributes are important in a way that can be especially powerful in low-data regimes. Similar structural kernels exist for various structured objects and, via COWBOYS, could now be used for molecular graph optimisation with graph kernels \cite{vishwanathan2010graph}, engineering design with 3d mesh kernels \cite{perez2024gaussian}, optimising computer code via tree kernels \cite{beck2015learning}, or protein design with new protein kernels \cite{groth2024kermut}. By enabling the direct use of such kernels, COWBOYS allows practitioners to incorporate a wealth of domain-specific prior knowledge. We hope our work helps revitalize interest in harnessing these powerful kernels across a range of complex design problems and spurs further synergy between rich generative models and carefully structured discriminative models. 

\textbf{Methodological extensions}. Extending COWBOYS to the important task of large-scale molecular optimisation raises several exciting avenues for future research. First, we need to develop methods for fitting sparse GPs to discrete molecular data—an inherently more challenging task--- where inducing points are more difficult to optimise, perhaps by applying techniques from \citet{burt2019rates} or \citet{chang2022fantasizing}. Second, integrating COWBOYS with LSBO techniques that fine-tune their VAEs during optimisation may become essential for performance once experimental budgets grows and VAE adaptation becomes more feasible. Finally, more sophisticated sampling algorithms could further improve MCMC efficiency (as is important for controlling the cost of COWBOYS under expensive VAE architectures) or allow us to exploit the full stochastic nature of the decoder, rather than restricting the search to the most likely decoding. Also, note that the condition in \ref{eq:opt_goal} does resemble the well-known Probability of Improvement (PI) acquisition function. While PI is intuitive, it is known to be suboptimal and can exhibit pathological behaviour and so adapting more sophisticated acquisition schemes into a COWBOYS framework is an exciting direction for future research, e.g. light-weight information-theoretic schemes for multi-fidelity \cite{moss2020mumbo}, multi-objective \cite{qing2023pf} and tackling more ambitious problems like quantile optimisation \cite{picheny2022bayesian}.

\section*{Acknowledgments}
This research was supported by AstraZeneca, Schmidt Sciences, and Lancaster University's  Mathematics for AI in Real-world Systems E3 grant.

\section*{Impact Statement}
Our work introduces an efficient framework for optimising over discrete sequences, a problem of growing interest in machine learning. Real-world benefits in molecule design include significantly reducing wet-lab resources. However, the same method could be misused in non-therapeutic applications. Thus, while optimisation is automatic, humans must decide how these results are applied, following appropriate ethical guidelines.

\bibliography{references}

\begin{thebibliography}{75}
\providecommand{\natexlab}[1]{#1}
\providecommand{\url}[1]{\texttt{#1}}
\expandafter\ifx\csname urlstyle\endcsname\relax
  \providecommand{\doi}[1]{doi: #1}\else
  \providecommand{\doi}{doi: \begingroup \urlstyle{rm}\Url}\fi

\bibitem[Andrieu \& Thoms(2008)Andrieu and Thoms]{andrieu2008tutorial}
Andrieu, C. and Thoms, J.
\newblock A tutorial on adaptive mcmc.
\newblock \emph{Statistics and computing}, 18:\penalty0 343--373, 2008.

\bibitem[Auer(2000)]{auer2000using}
Auer, P.
\newblock Using upper confidence bounds for online learning.
\newblock In \emph{Proceedings 41st annual symposium on foundations of computer science}, pp.\  270--279. IEEE, 2000.

\bibitem[Balandat et~al.(2020)Balandat, Karrer, Jiang, Daulton, Letham, Wilson, and Bakshy]{balandat2020botorch}
Balandat, M., Karrer, B., Jiang, D., Daulton, S., Letham, B., Wilson, A.~G., and Bakshy, E.
\newblock Botorch: A framework for efficient monte-carlo bayesian optimization.
\newblock \emph{Advances in neural information processing systems}, 33:\penalty0 21524--21538, 2020.

\bibitem[Beck et~al.(2015)Beck, Cohn, Hardmeier, and Specia]{beck2015learning}
Beck, D., Cohn, T., Hardmeier, C., and Specia, L.
\newblock Learning structural kernels for natural language processing.
\newblock \emph{Transactions of the Association for Computational Linguistics}, 2015.

\bibitem[Bergstra et~al.(2011)Bergstra, Bardenet, Bengio, and K{\'e}gl]{bergstra2011algorithms}
Bergstra, J., Bardenet, R., Bengio, Y., and K{\'e}gl, B.
\newblock Algorithms for hyper-parameter optimization.
\newblock \emph{Advances in neural information processing systems}, 24, 2011.

\bibitem[Bodin et~al.(2024)Bodin, Moss, and Ek]{bodin2024linear}
Bodin, E., Moss, H., and Ek, C.~H.
\newblock Linear combinations of latents in diffusion models: interpolation and beyond.
\newblock \emph{arXiv e-prints}, pp.\  arXiv--2408, 2024.

\bibitem[Brooks et~al.(2011)Brooks, Gelman, Jones, and Meng]{brooks2011handbook}
Brooks, S., Gelman, A., Jones, G., and Meng, X.-L.
\newblock \emph{Handbook of markov chain monte carlo}.
\newblock CRC press, 2011.

\bibitem[Brown et~al.(2019)Brown, Fiscato, Segler, and Vaucher]{brown2019guacamol}
Brown, N., Fiscato, M., Segler, M.~H., and Vaucher, A.~C.
\newblock Guacamol: benchmarking models for de novo molecular design.
\newblock \emph{Journal of chemical information and modeling}, 2019.

\bibitem[Burt et~al.(2019)Burt, Rasmussen, and Van Der~Wilk]{burt2019rates}
Burt, D., Rasmussen, C.~E., and Van Der~Wilk, M.
\newblock Rates of convergence for sparse variational gaussian process regression.
\newblock In \emph{International Conference on Machine Learning}, 2019.

\bibitem[Chang et~al.(2022)Chang, Verma, John, Picheny, Moss, and Solin]{chang2022fantasizing}
Chang, P.~E., Verma, P., John, S., Picheny, V., Moss, H., and Solin, A.
\newblock Fantasizing with dual gps in bayesian optimization and active learning.
\newblock \emph{arXiv preprint arXiv:2211.01053}, 2022.

\bibitem[Chen et~al.(2024)Chen, Duan, Li, Qi, Shi, and Gao]{chen2024pg}
Chen, T., Duan, Y., Li, D., Qi, L., Shi, Y., and Gao, Y.
\newblock Pg-lbo: Enhancing high-dimensional bayesian optimization with pseudo-label and gaussian process guidance.
\newblock In \emph{Proceedings of the AAAI Conference on Artificial Intelligence}, 2024.

\bibitem[Chu et~al.(2024)Chu, Park, Lee, and Kim]{chuinversion}
Chu, J., Park, J., Lee, S., and Kim, H.~J.
\newblock Inversion-based latent {B}ayesian optimization.
\newblock In \emph{The Thirty-eighth Annual Conference on Neural Information Processing Systems}, 2024.

\bibitem[Cotter et~al.(2013)Cotter, Roberts, Stuart, and White]{cotter2013mcmc}
Cotter, S., Roberts, G., Stuart, A., and White, D.
\newblock Mcmc methods for functions: Modifying old algorithms to make them faster.
\newblock \emph{Statistical Science}, 2013.

\bibitem[Daulton et~al.(2022)Daulton, Wan, Eriksson, Balandat, Osborne, and Bakshy]{daulton2022bayesian}
Daulton, S., Wan, X., Eriksson, D., Balandat, M., Osborne, M.~A., and Bakshy, E.
\newblock Bayesian optimization over discrete and mixed spaces via probabilistic reparameterization.
\newblock \emph{Advances in Neural Information Processing Systems}, 2022.

\bibitem[Deshwal \& Doppa(2021)Deshwal and Doppa]{deshwal2021combining}
Deshwal, A. and Doppa, J.
\newblock Combining latent space and structured kernels for {B}ayesian optimization over combinatorial spaces.
\newblock \emph{Advances in Neural Information Processing Systems}, 34:\penalty0 8185--8200, 2021.

\bibitem[Eissman et~al.(2018)Eissman, Levy, Shu, Bartzsch, and Ermon]{eissman2018bayesian}
Eissman, S., Levy, D., Shu, R., Bartzsch, S., and Ermon, S.
\newblock Bayesian optimization and attribute adjustment.
\newblock In \emph{Proc. 34th Conference on Uncertainty in Artificial Intelligence}, 2018.

\bibitem[Eriksson et~al.(2019)Eriksson, Pearce, Gardner, Turner, and Poloczek]{eriksson2019scalable}
Eriksson, D., Pearce, M., Gardner, J., Turner, R.~D., and Poloczek, M.
\newblock Scalable global optimization via local bayesian optimization.
\newblock \emph{Advances in neural information processing systems}, 32, 2019.

\bibitem[Gao et~al.(2022)Gao, Fu, Sun, and Coley]{gao2022sample}
Gao, W., Fu, T., Sun, J., and Coley, C.
\newblock Sample efficiency matters: a benchmark for practical molecular optimization.
\newblock \emph{Advances in neural information processing systems}, 2022.

\bibitem[Gardner et~al.(2018)Gardner, Pleiss, Weinberger, Bindel, and Wilson]{gardner2018gpytorch}
Gardner, J., Pleiss, G., Weinberger, K.~Q., Bindel, D., and Wilson, A.~G.
\newblock Gpytorch: Blackbox matrix-matrix gaussian process inference with gpu acceleration.
\newblock \emph{Advances in neural information processing systems}, 31, 2018.

\bibitem[G{\'o}mez-Bombarelli et~al.(2018)G{\'o}mez-Bombarelli, Wei, Duvenaud, Hern{\'a}ndez-Lobato, S{\'a}nchez-Lengeling, Sheberla, Aguilera-Iparraguirre, Hirzel, Adams, and Aspuru-Guzik]{gomez2018automatic}
G{\'o}mez-Bombarelli, R., Wei, J.~N., Duvenaud, D., Hern{\'a}ndez-Lobato, J.~M., S{\'a}nchez-Lengeling, B., Sheberla, D., Aguilera-Iparraguirre, J., Hirzel, T.~D., Adams, R.~P., and Aspuru-Guzik, A.
\newblock Automatic chemical design using a data-driven continuous representation of molecules.
\newblock \emph{ACS central science}, 2018.

\bibitem[Gonz{\'a}lez-Duque et~al.(2024)Gonz{\'a}lez-Duque, Michael, Bartels, Zainchkovskyy, Hauberg, and Boomsma]{gonzalez2024survey}
Gonz{\'a}lez-Duque, M., Michael, R., Bartels, S., Zainchkovskyy, Y., Hauberg, S., and Boomsma, W.
\newblock A survey and benchmark of high-dimensional bayesian optimization of discrete sequences.
\newblock \emph{arXiv preprint arXiv:2406.04739}, 2024.

\bibitem[Griffiths \& Hern{\'a}ndez-Lobato(2020)Griffiths and Hern{\'a}ndez-Lobato]{griffiths2020constrained}
Griffiths, R.-R. and Hern{\'a}ndez-Lobato, J.~M.
\newblock Constrained {B}ayesian optimization for automatic chemical design using variational autoencoders.
\newblock \emph{Chemical science}, 2020.

\bibitem[Griffiths et~al.(2022)Griffiths, Greenfield, Thawani, Jamasb, Moss, Bourached, Jones, McCorkindale, Aldrick, Fuchter, et~al.]{griffiths2022data}
Griffiths, R.-R., Greenfield, J.~L., Thawani, A.~R., Jamasb, A.~R., Moss, H.~B., Bourached, A., Jones, P., McCorkindale, W., Aldrick, A.~A., Fuchter, M.~J., et~al.
\newblock Data-driven discovery of molecular photoswitches with multioutput gaussian processes.
\newblock \emph{Chemical Science}, 2022.

\bibitem[Griffiths et~al.(2024)Griffiths, Klarner, Moss, Ravuri, Truong, Du, Stanton, Tom, Rankovic, Jamasb, et~al.]{griffiths2024gauche}
Griffiths, R.-R., Klarner, L., Moss, H., Ravuri, A., Truong, S., Du, Y., Stanton, S., Tom, G., Rankovic, B., Jamasb, A., et~al.
\newblock {GAUCHE}: a library for {G}aussian processes in chemistry.
\newblock \emph{Advances in Neural Information Processing Systems}, 36, 2024.

\bibitem[Grosnit et~al.(2021)Grosnit, Tutunov, Maraval, Griffiths, Cowen-Rivers, Yang, Zhu, Lyu, Chen, Wang, et~al.]{grosnit2021high}
Grosnit, A., Tutunov, R., Maraval, A.~M., Griffiths, R.-R., Cowen-Rivers, A.~I., Yang, L., Zhu, L., Lyu, W., Chen, Z., Wang, J., et~al.
\newblock High-dimensional {B}ayesian optimisation with variational autoencoders and deep metric learning.
\newblock \emph{arXiv preprint arXiv:2106.03609}, 2021.

\bibitem[Groth et~al.(2024)Groth, Kerrn, Olsen, Salomon, and Boomsma]{groth2024kermut}
Groth, P.~M., Kerrn, M., Olsen, L., Salomon, J., and Boomsma, W.
\newblock Kermut: Composite kernel regression for protein variant effects.
\newblock \emph{Advances in Neural Information Processing Systems}, 37, 2024.

\bibitem[Hairer et~al.(2014)Hairer, Stuart, and Vollmer]{hairer2014spectral}
Hairer, M., Stuart, A., and Vollmer, S.
\newblock Spectral gaps for a {M}etropolis--{H}astings algorithm in infinite dimensions.
\newblock \emph{The Annals of Applied Probability}, 2014.

\bibitem[Hennig \& Schuler(2012)Hennig and Schuler]{hennig2012entropy}
Hennig, P. and Schuler, C.~J.
\newblock Entropy search for information-efficient global optimization.
\newblock \emph{Journal of Machine Learning Research}, 2012.

\bibitem[Hensman et~al.(2013)Hensman, Fusi, and Lawrence]{hensman2013gaussian}
Hensman, J., Fusi, N., and Lawrence, N.~D.
\newblock Gaussian processes for big data.
\newblock In \emph{Proceedings of the Twenty-Ninth Conference on Uncertainty in Artificial Intelligence}, pp.\  282--290, 2013.

\bibitem[Hvarfner et~al.(2022)Hvarfner, Stoll, Souza, Lindauer, Hutter, and Nardi]{hvarfner2022pibo}
Hvarfner, C., Stoll, D., Souza, A., Lindauer, M., Hutter, F., and Nardi, L.
\newblock $\pi$bo: Augmenting acquisition functions with user beliefs for bayesian optimization.
\newblock In \emph{Tenth International Conference of Learning Representations, ICLR 2022}, 2022.

\bibitem[Hvarfner et~al.(2024)Hvarfner, Hellsten, and Nardi]{hvarfner2024vanilla}
Hvarfner, C., Hellsten, E.~O., and Nardi, L.
\newblock Vanilla bayesian optimization performs great in high dimension.
\newblock \emph{arXiv preprint arXiv:2402.02229}, 2024.

\bibitem[Jensen(2019)]{jensen2019graph}
Jensen, J.~H.
\newblock A graph-based genetic algorithm and generative model/monte carlo tree search for the exploration of chemical space.
\newblock \emph{Chemical science}, 10\penalty0 (12):\penalty0 3567--3572, 2019.

\bibitem[Jin et~al.(2018)Jin, Barzilay, and Jaakkola]{jin2018junction}
Jin, W., Barzilay, R., and Jaakkola, T.
\newblock Junction tree variational autoencoder for molecular graph generation.
\newblock In \emph{International conference on machine learning}, 2018.

\bibitem[Kingma \& Welling(2014)Kingma and Welling]{kingma2014auto}
Kingma, D.~P. and Welling, M.
\newblock Auto-encoding variational {B}ayes.
\newblock In \emph{Proceedings of the International Conference on Learning Representations (ICLR)}, 2014.

\bibitem[Kirschner et~al.(2019)Kirschner, Mutny, Hiller, Ischebeck, and Krause]{kirschner2019adaptive}
Kirschner, J., Mutny, M., Hiller, N., Ischebeck, R., and Krause, A.
\newblock Adaptive and safe bayesian optimization in high dimensions via one-dimensional subspaces.
\newblock In \emph{International Conference on Machine Learning}, pp.\  3429--3438. PMLR, 2019.

\bibitem[Kriege et~al.(2020)Kriege, Johansson, and Morris]{kriege2020survey}
Kriege, N.~M., Johansson, F.~D., and Morris, C.
\newblock A survey on graph kernels.
\newblock \emph{Applied Network Science}, 2020.

\bibitem[Kusner et~al.(2017)Kusner, Paige, and Hern{\'a}ndez-Lobato]{kusner2017grammar}
Kusner, M.~J., Paige, B., and Hern{\'a}ndez-Lobato, J.~M.
\newblock Grammar variational autoencoder.
\newblock In \emph{International conference on machine learning}, pp.\  1945--1954. PMLR, 2017.

\bibitem[Landrum(2013)]{landrum2013rdkit}
Landrum, G.
\newblock Rdkit documentation.
\newblock \emph{Release}, 2013.

\bibitem[Lee et~al.(2023)Lee, Chu, Kim, Ko, and Kim]{lee2023advancing}
Lee, S., Chu, J., Kim, S., Ko, J., and Kim, H.~J.
\newblock Advancing {B}ayesian optimization via learning correlated latent space.
\newblock In \emph{Thirty-seventh Conference on Neural Information Processing Systems}, 2023.

\bibitem[Lee et~al.(2024)Lee, Chu, Kim, Ko, and Kim]{lee2024advancing}
Lee, S., Chu, J., Kim, S., Ko, J., and Kim, H.~J.
\newblock Advancing {B}ayesian optimization via learning correlated latent space.
\newblock \emph{Advances in Neural Information Processing Systems}, 2024.

\bibitem[Lee et~al.(2025)Lee, Park, Chu, Yoon, and Kim]{leelatent}
Lee, S., Park, J., Chu, J., Yoon, M., and Kim, H.~J.
\newblock Latent bayesian optimization via autoregressive normalizing flows.
\newblock In \emph{The Thirteenth International Conference on Learning Representations}, 2025.

\bibitem[Maus et~al.(2022{\natexlab{a}})Maus, Jones, Moore, Kusner, Bradshaw, and Gardner]{maus2022local}
Maus, N., Jones, H., Moore, J., Kusner, M.~J., Bradshaw, J., and Gardner, J.
\newblock Local latent space {B}ayesian optimization over structured inputs.
\newblock \emph{Advances in Neural Information Processing Systems}, 2022{\natexlab{a}}.

\bibitem[Maus et~al.(2022{\natexlab{b}})Maus, Wu, Eriksson, and Gardner]{maus2022discovering}
Maus, N., Wu, K., Eriksson, D., and Gardner, J.
\newblock Discovering many diverse solutions with {B}ayesian optimization.
\newblock \emph{arXiv preprint arXiv:2210.10953}, 2022{\natexlab{b}}.

\bibitem[Maus et~al.(2024)Maus, Kim, Pleiss, Eriksson, Cunningham, and Gardner]{maus2024approximation}
Maus, N., Kim, K., Pleiss, G., Eriksson, D., Cunningham, J.~P., and Gardner, J.~R.
\newblock Approximation-aware bayesian optimization.
\newblock \emph{arXiv preprint arXiv:2406.04308}, 2024.

\bibitem[McHutchon \& Rasmussen(2011)McHutchon and Rasmussen]{mchutchon2011gaussian}
McHutchon, A. and Rasmussen, C.
\newblock Gaussian process training with input noise.
\newblock \emph{Advances in neural information processing systems}, 2011.

\bibitem[Mockus(2005)]{mockus2005bayesian}
Mockus, J.
\newblock The {B}ayesian approach to global optimization.
\newblock In \emph{System Modeling and Optimization: Proceedings of the 10th IFIP Conference New York City, USA, August 31--September 4, 1981}, pp.\  473--481. Springer, 2005.

\bibitem[Moriconi et~al.(2020)Moriconi, Deisenroth, and Sesh~Kumar]{moriconi2020high}
Moriconi, R., Deisenroth, M.~P., and Sesh~Kumar, K.
\newblock High-dimensional {B}ayesian optimization using low-dimensional feature spaces.
\newblock \emph{Machine Learning}, 2020.

\bibitem[Moss et~al.(2020{\natexlab{a}})Moss, Leslie, Beck, Gonzalez, and Rayson]{moss2020boss}
Moss, H., Leslie, D., Beck, D., Gonzalez, J., and Rayson, P.
\newblock Boss: {B}ayesian optimization over string spaces.
\newblock \emph{Advances in neural information processing systems}, 2020{\natexlab{a}}.

\bibitem[Moss \& Griffiths(2020)Moss and Griffiths]{moss2020gaussian}
Moss, H.~B. and Griffiths, R.-R.
\newblock Gaussian process molecule property prediction with flowmo.
\newblock \emph{arXiv preprint arXiv:2010.01118}, 2020.

\bibitem[Moss et~al.(2020{\natexlab{b}})Moss, Leslie, and Rayson]{moss2020mumbo}
Moss, H.~B., Leslie, D.~S., and Rayson, P.
\newblock Mumbo: Multi-task max-value bayesian optimization.
\newblock In \emph{Joint European Conference on Machine Learning and Knowledge Discovery in Databases}, 2020{\natexlab{b}}.

\bibitem[Moss et~al.(2021)Moss, Leslie, Gonzalez, and Rayson]{moss2021gibbon}
Moss, H.~B., Leslie, D.~S., Gonzalez, J., and Rayson, P.
\newblock Gibbon: General-purpose information-based bayesian optimisation.
\newblock \emph{Journal of Machine Learning Research}, 2021.

\bibitem[Moss et~al.(2022)Moss, Ober, and Picheny]{moss2022information}
Moss, H.~B., Ober, S.~W., and Picheny, V.
\newblock Information-theoretic inducing point placement for high-throughput bayesian optimisation.
\newblock \emph{arXiv preprint arXiv:2206.02437}, 2022.

\bibitem[Moss et~al.(2023)Moss, Ober, and Picheny]{moss2023inducing}
Moss, H.~B., Ober, S.~W., and Picheny, V.
\newblock Inducing point allocation for sparse gaussian processes in high-throughput bayesian optimisation.
\newblock In \emph{International Conference on Artificial Intelligence and Statistics}, 2023.

\bibitem[Notin et~al.(2021)Notin, Hern{\'a}ndez-Lobato, and Gal]{notin2021improving}
Notin, P., Hern{\'a}ndez-Lobato, J.~M., and Gal, Y.
\newblock Improving black-box optimization in vae latent space using decoder uncertainty.
\newblock \emph{Advances in Neural Information Processing Systems}, 2021.

\bibitem[Ober et~al.(2021)Ober, Rasmussen, and van~der Wilk]{ober2021promises}
Ober, S.~W., Rasmussen, C.~E., and van~der Wilk, M.
\newblock The promises and pitfalls of deep kernel learning.
\newblock In \emph{Uncertainty in Artificial Intelligence}, 2021.

\bibitem[Ober et~al.(2025)Ober, Power, Diethe, and Moss]{ober2025big}
Ober, S.~W., Power, S., Diethe, T., and Moss, H.~B.
\newblock Big batch {B}ayesian active learning by considering predictive probabilities.
\newblock \emph{arXiv preprint arXiv:2501.08223}, 2025.

\bibitem[Papenmeier et~al.(2023)Papenmeier, Nardi, and Poloczek]{papenmeier2023bounce}
Papenmeier, L., Nardi, L., and Poloczek, M.
\newblock Bounce: reliable high-dimensional bayesian optimization for combinatorial and mixed spaces.
\newblock \emph{Advances in Neural Information Processing Systems}, 2023.

\bibitem[Perez et~al.(2024)Perez, Da~Veiga, Garnier, and Staber]{perez2024gaussian}
Perez, R.~C., Da~Veiga, S., Garnier, J., and Staber, B.
\newblock Gaussian process regression with sliced wasserstein weisfeiler-lehman graph kernels.
\newblock In \emph{International Conference on Artificial Intelligence and Statistics}, 2024.

\bibitem[Picheny et~al.(2022)Picheny, Moss, Torossian, and Durrande]{picheny2022bayesian}
Picheny, V., Moss, H., Torossian, L., and Durrande, N.
\newblock Bayesian quantile and expectile optimisation.
\newblock In \emph{Uncertainty in Artificial Intelligence}, pp.\  1623--1633. PMLR, 2022.

\bibitem[Qing et~al.(2023{\natexlab{a}})Qing, Couckuyt, and Dhaene]{qing2023robust}
Qing, J., Couckuyt, I., and Dhaene, T.
\newblock A robust multi-objective {B}ayesian optimization framework considering input uncertainty.
\newblock \emph{Journal of Global Optimization}, 2023{\natexlab{a}}.

\bibitem[Qing et~al.(2023{\natexlab{b}})Qing, Moss, Dhaene, and Couckuyt]{qing2023pf}
Qing, J., Moss, H.~B., Dhaene, T., and Couckuyt, I.
\newblock Pf2es: Parallel feasible pareto frontier entropy search for multi-objective bayesian optimization.
\newblock In \emph{International Conference on Artificial Intelligence and Statistics}, 2023{\natexlab{b}}.

\bibitem[Ramchandran et~al.(2025)Ramchandran, Haussmann, and L{\"a}hdesm{\"a}ki]{ramchandranhigh}
Ramchandran, S., Haussmann, M., and L{\"a}hdesm{\"a}ki, H.
\newblock High-dimensional bayesian optimisation with gaussian process prior variational autoencoders.
\newblock In \emph{The Thirteenth International Conference on Learning Representations}, 2025.

\bibitem[Rasmussen(2003)]{rasmussen2003gaussian}
Rasmussen, C.~E.
\newblock Gaussian processes in machine learning.
\newblock In \emph{Summer school on machine learning}, pp.\  63--71. Springer, 2003.

\bibitem[Rezende et~al.(2014)Rezende, Mohamed, and Wierstra]{rezende2014stochastic}
Rezende, D.~J., Mohamed, S., and Wierstra, D.
\newblock Stochastic backpropagation and approximate inference in deep generative models.
\newblock In \emph{Proceedings of the 31st International Conference on Machine Learning}, 2014.

\bibitem[Schraudolph et~al.(2010)Schraudolph, Kondor, and Borgwardt]{vishwanathan2010graph}
Schraudolph, N.~N., Kondor, R., and Borgwardt, K.~M.
\newblock Graph kernels.
\newblock \emph{The Journal of Machine Learning Research}, 2010.

\bibitem[Shoemake(1985)]{shoemake1985animating}
Shoemake, K.
\newblock Animating rotation with quaternion curves.
\newblock In \emph{Proceedings of the 12th annual conference on Computer graphics and interactive techniques}, pp.\  245--254, 1985.

\bibitem[Song et~al.(2020{\natexlab{a}})Song, Meng, and Ermon]{song2020denoising}
Song, J., Meng, C., and Ermon, S.
\newblock Denoising diffusion implicit models.
\newblock \emph{arXiv preprint arXiv:2010.02502}, 2020{\natexlab{a}}.

\bibitem[Song et~al.(2020{\natexlab{b}})Song, Sohl-Dickstein, Kingma, Kumar, Ermon, and Poole]{song2020score}
Song, Y., Sohl-Dickstein, J., Kingma, D.~P., Kumar, A., Ermon, S., and Poole, B.
\newblock Score-based generative modeling through stochastic differential equations.
\newblock \emph{arXiv preprint arXiv:2011.13456}, 2020{\natexlab{b}}.

\bibitem[Stanton et~al.(2022)Stanton, Maddox, Gruver, Maffettone, Delaney, Greenside, and Wilson]{stanton2022accelerating}
Stanton, S., Maddox, W., Gruver, N., Maffettone, P., Delaney, E., Greenside, P., and Wilson, A.~G.
\newblock Accelerating {B}ayesian optimization for biological sequence design with denoising autoencoders.
\newblock In \emph{International Conference on Machine Learning}, 2022.

\bibitem[Tripp \& Hern{\'a}ndez-Lobato(2024)Tripp and Hern{\'a}ndez-Lobato]{tripp2024diagnosing}
Tripp, A. and Hern{\'a}ndez-Lobato, J.~M.
\newblock Diagnosing and fixing common problems in bayesian optimization for molecule design.
\newblock \emph{arXiv e-prints}, pp.\  arXiv--2406, 2024.

\bibitem[Tripp et~al.(2020)Tripp, Daxberger, and Hern{\'a}ndez-Lobato]{tripp2020sample}
Tripp, A., Daxberger, E., and Hern{\'a}ndez-Lobato, J.~M.
\newblock Sample-efficient optimization in the latent space of deep generative models via weighted retraining.
\newblock \emph{Advances in Neural Information Processing Systems}, 2020.

\bibitem[Tripp et~al.(2023)Tripp, Bacallado, Singh, and Hern\'{a}ndez-Lobato]{tripp2023tanimoto}
Tripp, A., Bacallado, S., Singh, S., and Hern\'{a}ndez-Lobato, J.~M.
\newblock Tanimoto random features for scalable molecular machine learning.
\newblock In \emph{Advances in Neural Information Processing Systems}, 2023.

\bibitem[Vakili et~al.(2021)Vakili, Moss, Artemev, Dutordoir, and Picheny]{vakili2021scalable}
Vakili, S., Moss, H., Artemev, A., Dutordoir, V., and Picheny, V.
\newblock Scalable thompson sampling using sparse gaussian process models.
\newblock \emph{Advances in neural information processing systems}, 34:\penalty0 5631--5643, 2021.

\bibitem[Vershynin(2018)]{vershynin2018high}
Vershynin, R.
\newblock \emph{High-dimensional probability: An introduction with applications in data science}, volume~47.
\newblock Cambridge university press, 2018.

\bibitem[White(2016)]{white2016sampling}
White, T.
\newblock Sampling generative networks.
\newblock \emph{arXiv preprint arXiv:1609.04468}, 2016.

\end{thebibliography}
\bibliographystyle{icml2025}

\newpage
\appendix
\onecolumn

\newpage
\section{COWBOYS' sampler details}
\label{appendix:sampling}

We now provide full details for our PCN MCMC sampler \citep{cotter2013mcmc} as Algorithms \ref{alg:COWBOYS_SAMPLE} and \ref{alg:PCN} which we use to get a set of samples from COWBOYS sampling objective (\ref{eq:cowboys_z}). Note that we keep track of the latent value that decoded to give the best structure so far $\bm z_{best}$ and use this to start our chains. We also employ an adaptive choice of PCN's scale parameter $\beta$ following \citet{andrieu2008tutorial}. Often, especially early in the optimisation before the region satisfying likely to satisfy $f(\bm x)>f^*$ narrows, our PCN sampler can return a larger number of samples than our desired batch size. In this case, we select the $B$ samples that achieve the highest expected utility under greedy maximisation of the batch Expected Improvement (qEI) acquisition function provided in \texttt{BoTorch} \citep{balandat2020botorch}. If very large batches are required, then a subset could be chosen using Thompson  following \citet{vakili2021scalable}. Although this subset selection strategy breaks MCMC’s detailed balance, it ensures that we return a diverse, high-value subset of samples to COWBOYS -- samples that are likely to be approximate anyway due to relatively small number of MCMC iterations and lack of burn-in phase. Note that in practical implementations, our PCN (Algorithm \ref{alg:PCN}) is trivially parallelised by running all chains present in the outer loop at once.

\begin{algorithm}[]
   \caption{\label{alg:COWBOYS_SAMPLE}COWBOYS' MCMC Sampler}
\begin{algorithmic}
   \STATE {\bfseries Input:} Batch size $B$, number chains $C$, number steps $S$, threshold $f^*$, starting latent code$\bm z_{best}$
    \STATE $Z_{samples}\leftarrow \textrm{PCN}(C, S, f^*, \bm z_{best})$ \COMMENT{Algorithm \ref{alg:PCN}}
    \WHILE[Keep sampling if needed]{$|Z_{samples}| < B$}
    \STATE $Z_{samples}\leftarrow Z_{samples}\bigcup\textrm{PCN}(C, S, f^*, \bm z_{best})$ \COMMENT{Algorithm \ref{alg:PCN} again}
    \ENDWHILE
    \IF{$|Z_{samples}|>B$}
    \STATE $Z_{samples} \leftarrow \argmax_{\substack{Z\subseteq Z_{samples} \\  : |Z|=B}} \textrm{qEI}(Z)$ \COMMENT{choose $B$ best according to batch utility}
    \ENDIF
    \STATE \textbf{return}   $Z_{chosen}$ 
\end{algorithmic}
\end{algorithm}

\begin{algorithm}[]
   \caption{\label{alg:PCN}PCN}
\begin{algorithmic}
   \STATE {\bfseries Input:} number chains $C$, number steps $S$, best score so far $f^*$, starting latent code$\bm z_{best}$
   \STATE $Z_{samples}\leftarrow \emptyset$
   \FOR{$c\in\{1,..,C\}$}
    \STATE $\beta\leftarrow 0.1$  \COMMENT{Initialise step size}
   \STATE $\bm z_{\textrm{current}}\leftarrow \bm z_{\textrm{best}}$ \COMMENT{Initialise new chain at best so far}
   \FOR[Do MCMC step]{$n\in\{1,..,S\}$}
    \STATE  $\bm z_{\textrm{proposal}}\sim\mathcal{N}_d(\sqrt{1-\beta^2}\bm z_{\textrm{current}}, \beta^2I_{d}) $\COMMENT{Sample from PCN proposal density}
    \STATE  $\alpha \leftarrow \min\left(1, \frac{p\bigl(g_{\theta,\bm z_{\textrm{proposal}}} > f^* \mid D^{\mathcal{X}}_n\bigr)\times p(\bm z_{\textrm{proposal}})}{p\bigl(g_{\theta,\bm z_{\textrm{current}}} > f^* \mid D^{\mathcal{X}}_n\bigr)\times p(\bm z_{\textrm{current}})}\right)$\COMMENT{Calculate acceptance probability using prior and likelihood from (\ref{eq:decomp})}
    \STATE $u \sim U[0,1]$
    \IF[Stochastic acceptance step]{$\alpha > u$}
        \STATE $\bm z_{\textrm{current}}\leftarrow\bm z_{\textrm{proposal}}$ 
        \STATE $Z_{samples}\cup\{z_{\textrm{current}}\}$ \COMMENT{Store sample}
    \ENDIF
    \STATE $\beta\leftarrow\beta + 0.1 (\alpha -0.243)$ \COMMENT{Adaptive MCMC update}
   \ENDFOR
   \ENDFOR
    \STATE \textbf{return} $Z_{samples}$ \COMMENT{return all accepted samples}
\end{algorithmic}
\end{algorithm}

\newpage
\section{COWBOYS' Ablation Study}
\label{appendix:ablate}
Table \ref{table:ablate} examines COWBOYS' robustness to different configurations of its PCN MCMC sampler across the benchmark problems of \citet{gonzalez2024survey}, demonstrating that COWBOYS is not sensitive to the number of MCMC chains or the number of MCMC steps, unless they are set to be very small. Of course,  no single parameter choice can yield uniformly optimal performance across different objective functions, owing to varying degrees of model mismatch between our GP and each specific problem objective and varying degrees of difficulty/locality of the optimisation problems. The final column of the table highlights the combined impact of COWBOYS' two primary components (its MCMC sampling, and its modelling in structure space) by demonstrating that if we apply COWBOYS-style sampling but with a latent-space GP model, we lose performance. More precisely, rather than fitting a Tanimoto GP in the structure space, we instead follow an approach closer to LSBO, and fit a latent space GP, i.e., we replace the structured GP model $\hat{f}$ (recall $\hat{f}(h_{\theta}(\bm z))$) with a direct GP model $\tilde{g}(\bm z)$, but otherwise proceed as in COWBOYS.

\begin{table*}
\tabcolsep=0.1cm
\begin{tabular}{l|lllll|l|}
\hline
Oracle                    & \begin{tabular}[c]{@{}l@{}}COWBOYS\\ 10 chains\\ 100 steps\end{tabular} & \begin{tabular}[c]{@{}l@{}}COWBOYS\\ 1 chains\\ 100 steps\end{tabular} & \begin{tabular}[c]{@{}l@{}}COWBOYS\\ 10 chains\\ 50 steps\end{tabular} & \begin{tabular}[c]{@{}l@{}}COWBOYS\\ 1 chains\\ 50 steps\end{tabular} & \begin{tabular}[c]{@{}l@{}}COWBOYS\\ 1 chains\\ 10 steps\end{tabular} & \begin{tabular}[c]{@{}l@{}}COWBOYS \\with Latent GP \\ 10 chains\\ 100 steps\end{tabular} \\ \hline
albuterol\_similarity     & \cellcolor[HTML]{67FD9A}0.454+/-0.06                                    & \cellcolor[HTML]{009901}0.472+/-0.08                                   & \cellcolor[HTML]{67FD9A}0.450+/-0.06                                   & \cellcolor[HTML]{67FD9A}0.435+/-0.05                                  & \cellcolor[HTML]{67FD9A}0.464+/-0.07                                  & \cellcolor[HTML]{67FD9A}0.459+/-0.09                                                          \\
amlodipine\_mpo           & \cellcolor[HTML]{67FD9A}0.461+/-0.02                                    & \cellcolor[HTML]{009901}0.477+/-0.04                                   & \cellcolor[HTML]{67FD9A}0.471+/-0.02                                   & \cellcolor[HTML]{FFFFFF}0.450+/-0.01                                  & \cellcolor[HTML]{67FD9A}0.461+/-0.02                                  & \cellcolor[HTML]{FFFFFF}0.457+/-0.01                                                          \\
celecoxib\_rediscovery    & \cellcolor[HTML]{009901}0.264+/-0.04                                    & \cellcolor[HTML]{FFFFFF}0.217+/-0.02                                   & \cellcolor[HTML]{67FD9A}0.240+/-0.03                                   & \cellcolor[HTML]{FFFFFF}0.224+/-0.02                                  & \cellcolor[HTML]{FFFFFF}0.240+/-0.04                                  & \cellcolor[HTML]{FFFFFF}0.202+/-0.00                                                          \\
deco\_hop                 & \cellcolor[HTML]{67FD9A}0.569+/-0.01                                    & \cellcolor[HTML]{67FD9A}0.570+/-0.01                                   & \cellcolor[HTML]{009901}0.572+/-0.02                                   & \cellcolor[HTML]{FFFFFF}0.562+/-0.01                                  & \cellcolor[HTML]{67FD9A}0.568+/-0.01                                  & \cellcolor[HTML]{FFFFFF}0.562+/-0.01                                                          \\
drd2\_docking             & \cellcolor[HTML]{67FD9A}0.214+/-0.23                                    & \cellcolor[HTML]{009901}0.342+/-0.17                                   & \cellcolor[HTML]{67FD9A}0.325+/-0.17                                   & \cellcolor[HTML]{67FD9A}0.217+/-0.21                                  & \cellcolor[HTML]{67FD9A}0.266+/-0.08                                  & \cellcolor[HTML]{67FD9A}0.241+/-0.26                                                          \\
fexofenadine\_mpo         & \cellcolor[HTML]{FFFFFF}0.671+/-0.02                                    & \cellcolor[HTML]{67FD9A}0.682+/-0.02                                   & \cellcolor[HTML]{67FD9A}0.681+/-0.02                                   & \cellcolor[HTML]{FFFFFF}0.677+/-0.01                                  & \cellcolor[HTML]{009901}0.698+/-0.02                                  & \cellcolor[HTML]{67FD9A}0.674+/-0.03                                                          \\
gsk3\_beta                & \cellcolor[HTML]{009901}0.410+/-0.02                                    & \cellcolor[HTML]{FFFFFF}0.368+/-0.03                                   & \cellcolor[HTML]{67FD9A}0.358+/-0.06                                   & \cellcolor[HTML]{67FD9A}0.354+/-0.10                                  & \cellcolor[HTML]{FFFFFF}0.322+/-0.05                                  & \cellcolor[HTML]{FFFFFF}0.290+/-0.03                                                          \\
isomer\_c7h8n2o2          & \cellcolor[HTML]{FFFFFF}0.934+/-0.05                                    & \cellcolor[HTML]{009901}1.000+/-0.00                                   & \cellcolor[HTML]{FFFFFF}0.889+/-0.06                                   & \cellcolor[HTML]{67FD9A}0.976+/-0.05                                  & \cellcolor[HTML]{FFFFFF}0.921+/-0.06                                  & \cellcolor[HTML]{FFFFFF}0.880+/-0.07                                                          \\
isomer\_c9h10n2o2pf2cl    & \cellcolor[HTML]{67FD9A}0.634+/-0.06                                    & \cellcolor[HTML]{009901}0.690+/-0.08                                   & \cellcolor[HTML]{67FD9A}0.653+/-0.05                                   & \cellcolor[HTML]{67FD9A}0.634+/-0.08                                  & \cellcolor[HTML]{FFFFFF}0.649+/-0.04                                  & \cellcolor[HTML]{FFFFFF}0.622+/-0.05                                                          \\
jnk3                      & \cellcolor[HTML]{67FD9A}0.138+/-0.01                                    & \cellcolor[HTML]{67FD9A}0.138+/-0.05                                   & \cellcolor[HTML]{FFFFFF}0.124+/-0.02                                   & \cellcolor[HTML]{009901}0.146+/-0.02                                  & \cellcolor[HTML]{67FD9A}0.144+/-0.01                                  & \cellcolor[HTML]{67FD9A}0.134+/-0.03                                                          \\
median\_1                 & \cellcolor[HTML]{67FD9A}0.211+/-0.02                                    & \cellcolor[HTML]{FFFFFF}0.198+/-0.02                                   & \cellcolor[HTML]{009901}0.218+/-0.02                                   & \cellcolor[HTML]{67FD9A}0.198+/-0.03                                  & \cellcolor[HTML]{FFFFFF}0.175+/-0.02                                  & \cellcolor[HTML]{FFFFFF}0.170+/-0.03                                                          \\
median\_2                 & \cellcolor[HTML]{67FD9A}0.156+/-0.01                                    & \cellcolor[HTML]{009901}0.162+/-0.01                                   & \cellcolor[HTML]{FFFFFF}0.150+/-0.01                                   & \cellcolor[HTML]{67FD9A}0.155+/-0.01                                  & \cellcolor[HTML]{FFFFFF}0.158+/-0.01                                  & \cellcolor[HTML]{FFFFFF}0.145+/-0.01                                                          \\
mestranol\_similarity     & \cellcolor[HTML]{67FD9A}0.423+/-0.04                                    & \cellcolor[HTML]{FFFFFF}0.384+/-0.01                                   & \cellcolor[HTML]{009901}0.458+/-0.05                                   & \cellcolor[HTML]{FFFFFF}0.380+/-0.03                                  & \cellcolor[HTML]{FFFFFF}0.388+/-0.03                                  & \cellcolor[HTML]{FFFFFF}0.362+/-0.04                                                          \\
osimetrinib\_mpo          & \cellcolor[HTML]{67FD9A}0.749+/-0.02                                    & \cellcolor[HTML]{FFFFFF}0.722+/-0.03                                   & \cellcolor[HTML]{FFFFFF}0.731+/-0.01                                   & \cellcolor[HTML]{FFFFFF}0.733+/-0.02                                  & \cellcolor[HTML]{009901}0.754+/-0.02                                  & \cellcolor[HTML]{FFFFFF}0.721+/-0.02                                                          \\
perindopril\_mpo          & \cellcolor[HTML]{9AFF99}0.390+/-0.03                                    & \cellcolor[HTML]{9AFF99}0.382+/-0.04                                   & \cellcolor[HTML]{9AFF99}0.388+/-0.03                                   & \cellcolor[HTML]{9AFF99}0.376+/-0.04                                  & \cellcolor[HTML]{009901}0.399+/-0.03                                  & \cellcolor[HTML]{9AFF99}0.357+/-0.06                                                          \\
ranolazine\_mpo           & \cellcolor[HTML]{009901}0.652+/-0.03                                    & \cellcolor[HTML]{67FD9A}0.648+/-0.04                                   & \cellcolor[HTML]{67FD9A}0.643+/-0.02                                   & \cellcolor[HTML]{67FD9A}0.638+/-0.03                                  & \cellcolor[HTML]{67FD9A}0.628+/-0.04                                  & \cellcolor[HTML]{FFFFFF}0.601+/-0.02                                                          \\
rdkit\_logp               & \cellcolor[HTML]{67FD9A}9.733+/-0.96                                    & \cellcolor[HTML]{67FD9A}10.678+/-0.99                                  & \cellcolor[HTML]{67FD9A}10.508+/-0.57                                  & \cellcolor[HTML]{009901}10.690+/-0.69                                 & \cellcolor[HTML]{67FD9A}9.517+/-0.53                                  & \cellcolor[HTML]{67FD9A}9.406+/-1.34                                                          \\
rdkit\_qed                & \cellcolor[HTML]{67FD9A}0.917+/-0.02                                    & \cellcolor[HTML]{67FD9A}0.909+/-0.03                                   & \cellcolor[HTML]{009901}0.926+/-0.01                                   & \cellcolor[HTML]{67FD9A}0.915+/-0.02                                  & \cellcolor[HTML]{67FD9A}0.906+/-0.03                                  & \cellcolor[HTML]{67FD9A}0.917+/-0.02                                                          \\
sa\_tdc                   & \cellcolor[HTML]{009901}8.047+/-0.15                                    & \cellcolor[HTML]{FFFFFF}7.909+/-0.07                                   & \cellcolor[HTML]{67FD9A}8.012+/-0.13                                   & \cellcolor[HTML]{FFFFFF}7.982+/-0.02                                  & \cellcolor[HTML]{FFFFFF}7.945+/-0.03                                  & \cellcolor[HTML]{FFFFFF}7.798+/-0.13                                                          \\
scaffold\_hop             & \cellcolor[HTML]{009901}0.451+/-0.02                                    & \cellcolor[HTML]{FFFFFF}0.435+/-0.01                                   & \cellcolor[HTML]{67FD9A}0.441+/-0.02                                   & \cellcolor[HTML]{67FD9A}0.444+/-0.02                                  & \cellcolor[HTML]{FFFFFF}0.440+/-0.01                                  & \cellcolor[HTML]{FFFFFF}0.427+/-0.01                                                          \\
sitagliptin\_mpo          & \cellcolor[HTML]{009901}0.388+/-0.05                                    & \cellcolor[HTML]{FFFFFF}0.259+/-0.07                                   & \cellcolor[HTML]{FFFFFF}0.277+/-0.06                                   & \cellcolor[HTML]{FFFFFF}0.246+/-0.03                                  & \cellcolor[HTML]{FFFFFF}0.247+/-0.06                                  & \cellcolor[HTML]{FFFFFF}0.276+/-0.09                                                          \\
thiothixene\_rediscovery  & \cellcolor[HTML]{67FD9A}0.243+/-0.02                                    & \cellcolor[HTML]{009901}0.247+/-0.03                                   & \cellcolor[HTML]{67FD9A}0.244+/-0.02                                   & \cellcolor[HTML]{67FD9A}0.238+/-0.01                                  & \cellcolor[HTML]{009901}0.247+/-0.02                                  & \cellcolor[HTML]{FFFFFF}0.225+/-0.01                                                          \\
troglitazone\_rediscovery & \cellcolor[HTML]{FFFFFF}0.197+/-0.01                                    & \cellcolor[HTML]{67FD9A}0.205+/-0.02                                   & \cellcolor[HTML]{67FD9A}0.208+/-0.02                                   & \cellcolor[HTML]{009901}0.214+/-0.02                                  & \cellcolor[HTML]{67FD9A}0.203+/-0.02                                  & \cellcolor[HTML]{FFFFFF}0.192+/-0.01                                                          \\
valsartan\_smarts         & \cellcolor[HTML]{009901}0.000+/-0.00                                    & \cellcolor[HTML]{009901}0.000+/-0.00                                   & \cellcolor[HTML]{009901}0.000+/-0.00                                   & \cellcolor[HTML]{009901}0.000+/-0.00                                  & \cellcolor[HTML]{009901}0.000+/-0.00                                  & \cellcolor[HTML]{009901}0.000+/-0.00                                                          \\
zaleplon\_mpo             & \cellcolor[HTML]{9AFF99}0.382+/-0.03                                    & \cellcolor[HTML]{9AFF99}0.379+/-0.05                                   & \cellcolor[HTML]{009901}0.396+/-0.01                                   & \cellcolor[HTML]{9AFF99}0.346+/-0.05                                  & \cellcolor[HTML]{9AFF99}0.357+/-0.04                                  & \cellcolor[HTML]{9AFF99}0.340+/-0.07                                                          \\ \hline
\end{tabular}
\caption{Average performance ($\pm$ s.d.) over 5 repetitions of COWBOYS on the discrete BO benchmarking suite of \cite{gonzalez2024survey}. We stress best average scores achieved after 300 evaluations (dark) and scores within a single standard deviation of best (light). All runs of COWBOYS achieve roughly comparable results except the two on the far right, i.e, using too few MCMC chains/iterations or modifying COWBOYS to use a latent space GP, respectively.\label{table:ablate}}
\end{table*}

In order to diagnose if our deterministic decoder approximation is limiting the flexibility of our proposed molecules, We also ran the MCMC as described (to get a chosen \textbf{z}) but then sampled from the full stochastic decoder to get the chosen molecule. This lead to a significant drop in performance for 10 of the 25 tasks (and returning statistically similar scores for the other tasks), which we hypothesise is due to the fact that the resulting “sampled” molecules are now not the same as the ones passed to the GP to see if the considered latent code will yield an improvement in score – i.e. we return to a problem similar to that of the alignment issues that we discuss in Section \ref{section:poor_model}.

Finally, we also investigated the robustness of COWBOYS against degraded surrogate mode performance (to mimic settings where it may be hard to propose a structure-space kernel). Here, we deliberately impaired the Tanimoto GP surrogate by randomly swapping ones to zeros in molecular fingerprints with varying probabilities. This experiment corresponds to adding significant input noise to the fingerprints by randomly removing the count of particular (hashed) molecular substructures. Our findings indicated that COWBOYS maintains robust performance except under the most severe perturbations. In summary, perturbing Tanimoto entries with probability 0.01, 0.1, and 0.5 lead to a statistically significant drop in the best molecule found across 0, 3, and 8 tasks from the 25 molecular optimization tasks, respectively.

\end{document}